\pdfoutput=1

\documentclass[11pt]{article}
\usepackage{color}
\usepackage{colortbl}
\usepackage[table, dvipsnames]{xcolor}
\definecolor{intnull}{RGB}{225, 245, 164}
\definecolor{intyellow}{RGB}{250, 244, 135}
\definecolor{intgray}{RGB}{233, 239, 240}
\definecolor{intred}{RGB}{245, 224, 201}
\definecolor{patterngreen}{RGB}{102, 212, 149}
\definecolor{semanticblue}{RGB}{152, 227, 226}

\usepackage{acl}

\usepackage{hyperref}       
\usepackage{url}            
\usepackage{booktabs}
\usepackage{amsfonts}       

\usepackage{times}
\usepackage{latexsym}
\usepackage{soul}
\usepackage{amsmath}
\usepackage{CJKutf8}
\usepackage{multirow}
\usepackage{threeparttable}
\usepackage{float}
\usepackage{enumitem}
\usepackage{arydshln}
\usepackage{caption}
\usepackage[most]{tcolorbox}
\usepackage{subcaption}
\usepackage{comment}
\usepackage{algorithm}
\usepackage{algpseudocode}
\usepackage{threeparttable}
\usepackage{amssymb}
\usepackage{bbm}
\usepackage{fullwidth}
\usepackage[T1]{fontenc}

\usepackage[utf8]{inputenc}

\usepackage{microtype}

\usepackage{inconsolata}

\usepackage{graphicx}
\usepackage[]{mdframed}
\usepackage{pifont}
\newcommand*\colourcheck[1]{%
  \expandafter\newcommand\csname #1check\endcsname{\textcolor{#1}{\ding{52}}}%
}
\colourcheck{blue}
\colourcheck{green}
\colourcheck{red}

\def\mystrut(#1,#2){\vrule height #1pt depth #2pt width 0pt}   

\def\arrvline{\hfil\kern\arraycolsep\vline\kern-\arraycolsep\hfilneg}

\definecolor{myblue}{rgb}{0.82, 0.94, 0.75}
\definecolor{mygreen}{rgb}{0.64, 0.76, 0.68}
\definecolor{myyellow}{rgb}{0.88, 0.54, 0.35}
\definecolor{mydarkgreen}{rgb}{0.68, 0.9, 0.8}
\definecolor{mypink}{rgb}{0.2, 0.87, 0.2}
\definecolor{purple}{rgb}{0.5,0,1}
\definecolor{dcyan}{rgb}{0.2,0.6,0.5}
\definecolor{light-gray}{gray}{0.95} 
\definecolor{darkgreen}{RGB}{0,140,0}
\definecolor{darkred}{RGB}{200,0,0}
\definecolor{darkredTwo}{RGB}{183,20,20}
\definecolor{darkblueTwo}{RGB}{10,10,135}
\definecolor{lightgreen}{RGB}{197, 237, 208}
\definecolor{lightred}{RGB}{255,205,212}
\definecolor{lightyellow}{RGB}{255,240,160}
\definecolor{lightblue}{RGB}{195,221,255}
\definecolor{lightpurple}{RGB}{232,209,255}
\definecolor{lightgray}{RGB}{205,205,205}

\newtcbox{\mybox}[1][red]{on line,
    colback=#1, colframe=#1, boxsep=0pt, boxrule=0pt, size=small, arc=1mm,fontupper=\color{white},right=0pt,left=0pt,boxsep=2pt}

\newcommand{\daniel}[1]{{\color{purple}[DK: #1]}}

\newcommand{\yining}[1]{{\color{ForestGreen}[YL: #1]}}

\newcommand\Tstrut{\rule{0pt}{2.6ex}}         
\newcommand\Bstrut{\rule[-0.9ex]{0pt}{0pt}}   

\newcommand{\gear}[0]{\includegraphics[width=0.2cm,trim=10cm 3cm 0cm 5cm]{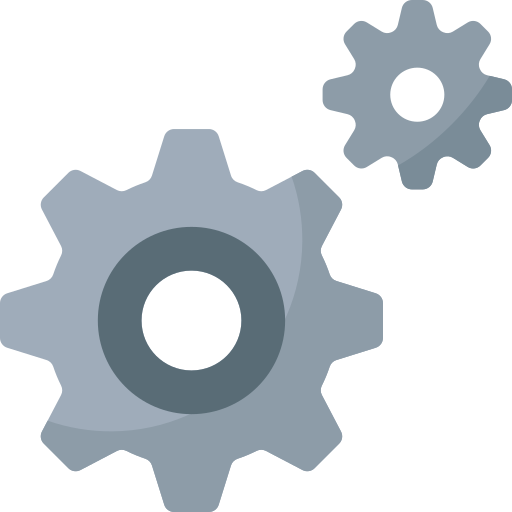}} 

\newcommand{\llm}{\textcolor{darkredTwo}{LLM}}
\newcommand{\slm}{\textcolor{darkblueTwo}{SLM}}

\newcommand{\name}{{\fontfamily{cmss}\selectfont GEAR}}

\newcommand{\calc}{{\tt Calculator}}
\newcommand{\mt}{{\tt MT}}
\newcommand{\qa}{{\tt QA}}
\newcommand{\wiki}{{\tt Wiki}}
\newcommand{\mlqa}{{\tt Multilingual QA}}
\newcommand{\tz}{{\tt Timezone Converter}}
\newcommand{\expcalc}{{\tt Exponential Calculator}}
\newcommand{\logcalc}{{\tt Logarithmic Calculator}}
\newcommand{\sleep}{{\tt Sleep}}
\newcommand{\map}{{\tt Movement Controller}}
\newcommand{\image}{{\tt Image Generation}}

\newcommand{\gptj}{{\tt \textcolor{darkredTwo}{GPT-J}}}
\newcommand{\gptjText}{{GPT-J}}

\newcommand{\gptDavinciThree}{{\tt \textcolor{darkredTwo}{GPT3$_{\text{davinci-003}}$}}}

\newcommand{\gptNeo}{{\tt \textcolor{darkblueTwo}{GPT-Neo}}}

\newcommand{\gptTwoLarge}{{\tt \textcolor{darkblueTwo}{GPT2$_{\text{large}}$}}}

\newcommand{\gptTwo}{{\tt \textcolor{darkblueTwo}{GPT2$_{\text{medium}}$}}}

\newcommand{\miniLM}{{\tt \textcolor{darkblueTwo}{MiniLM}}}

\newcommand{\gptThree}{{\tt \textcolor{darkredTwo}{GPT-3}}}
\newcommand{\gptThreeText}{{GPT-3}}

\newcommand{\mpnet}{{\tt \textcolor{darkblueTwo}{MPNet}}}

    \makeatletter
\def\@fnsymbol#1{\ensuremath{\ifcase#1\or 
 \heartsuit \or \ddagger\or
   \mathsection\or \mathparagraph\or \|\or **\or \dagger\dagger
   \or \ddagger\ddagger \else\@ctrerr\fi}}
    \makeatother

\title{\name: Efficient Tool Generalization Method for Augmented Language Model}
\title{\name: Efficient and Generalizable Tool Selection}
\title{\name: Augmented Language Models with \\ Efficient and Generalizable Tool Selection}
\title{\name: Language Models Augmented with \\ Efficient and Generalizable Tool Selection}
\title{\name: Language Models Augmented with \\ Efficient and Generalizable Tool Selection}
\title{\gear~\name: Generalizable and Efficient Augmented Tool Resolution}
\title{
\vspace*{-0.5in}
{{\small \hfill EACL 2024}\\
\vspace*{.25in}}
\gear~\name: Augmenting Language Models with \\ 
\hspace{0.2cm}Generalizable and Efficient Tool Resolution}

\author{Yining Lu\thanks{\:\;\;\;Equal contribution} \; \and Haoping Yu$^\heartsuit$ \and Daniel Khashabi \\
       Johns Hopkins University, Baltimore, MD \\ \texttt{\{ylu130, hyu90, danielk\}@jhu.edu}}

\usepackage{mleftright}
\newcommand{\setof}[1]{\mleft \{ #1 \mright\}}  %

\begin{document}
\maketitle
\begin{abstract}
Augmenting large language models (\llm{}) to use external tools enhances their performance across a variety of tasks. 
However, prior works over-rely on task-specific demonstration of tool use that limits their generalizability and computational cost due to making many calls to large-scale \llm{}s.
We introduce \name, a computationally efficient query-tool grounding algorithm that is generalizable to various tasks that require tool use while not relying on task-specific demonstrations.
\name{} achieves better efficiency by delegating tool grounding and execution to small language models (\slm{}) and \llm{}, respectively; while leveraging semantic and pattern-based evaluation at both question and answer levels for generalizable tool grounding. We evaluate \name{} on 14 datasets across 6 downstream tasks, demonstrating its strong generalizability to novel tasks, tools and different \slm{}s. Despite offering more efficiency, \name{} achieves higher precision in tool grounding compared to prior strategies using \llm{} prompting, thus improving downstream accuracy at a reduced computational cost. For example, we demonstrate that \name{}-augmented \gptjText{} and \gptThreeText{} outperform counterpart tool-augmented baselines because of better tool use.
\end{abstract}

\section{Introduction}
\newcommand{\arrowdownright}{\hspace{0.03cm}\reflectbox{\rotatebox[origin=c]{180}{$\Rsh$}}\hspace{0.03cm}}

\begin{figure}[ht]
    \centering
    \includegraphics[width=\linewidth,trim=0cm 0cm 0cm 1cm]{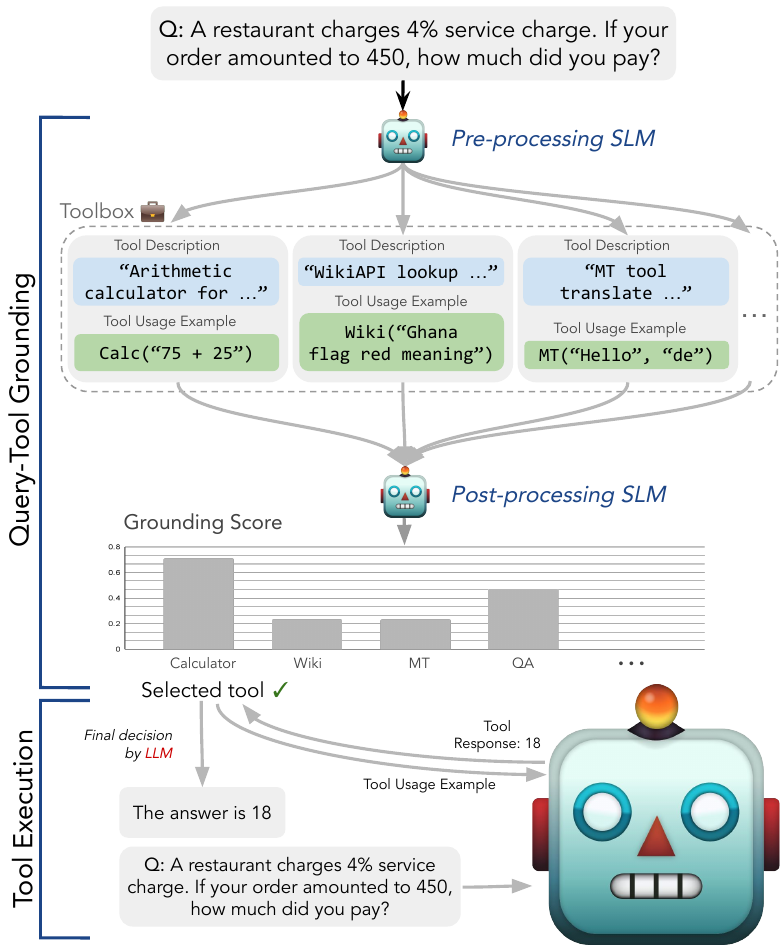}
    \caption{\name{} leverages small language models (\slm) to facilitate the process of \emph{tool grounding} for a given query and has the ability to add and utilize new tools for novel tasks without the need for fine-tuning or extra demonstrations. \name{} utilizes a large language model (\llm) in the \emph{tool execution} module to ensure the accuracy of the final answer.}
    \label{Teaser_Figure}
\end{figure}

\begin{table*}[t]
\footnotesize
\centering
\setlength{\tabcolsep}{2pt}
\resizebox{\textwidth}{!}{
\begin{tabular}{@{}lcccccc@{}}
\toprule
Feature                              & CoT                                                                                      & Zero-shot CoT                                                            & Toolformer                                                                    & ToolkenGPT                                                                                   & ART                                                                                                            & \name{}                                                 \\ \midrule
\rowcolor{intgray} Tool Use                             &   \textcolor{red}{\ding{55}}                                                                                       &  \textcolor{red}{\ding{55}}                                                                        & \greencheck                                                      & \greencheck                                                                                                                                           & \greencheck                                                                                       & \greencheck                                                 \\
Novel Task Generalization            &       \textcolor{red}{\ding{55}}                                                                                   & \greencheck                                                 &    \greencheck                                                                             &  \textcolor{red}{\ding{55}}                                                                                                                                                                         & \textcolor{red}{\ding{55}}                                                                                                               & \greencheck                                                 \\
\rowcolor{intgray}Extensibility to New Tools at Inference          &     \textit{N/A}                                                                                    &      \textit{N/A}                                            &             \textcolor{red}{\ding{55}}                                                                   &       \textcolor{red}{\ding{55}}                                                                                                                                                                     &   \greencheck                                                                                                             & \greencheck                                                 \\
\multirow{2}{*}{Grounding Algorithm} & \multicolumn{1}{c}{\multirow{2}{*}{\textit{N/A}}}                                                    & \multicolumn{1}{c}{\multirow{2}{*}{\textit{N/A}}}                                    & \multicolumn{1}{c}{\multirow{2}{*}{Finetune}}                                 & \multicolumn{1}{c}{\multirow{2}{*}{\begin{tabular}[c]{@{}c@{}}\llm{} \\ Generation\end{tabular}}}   & \multicolumn{1}{c}{\multirow{2}{*}{\begin{tabular}[c]{@{}c@{}}\llm{}-Based or \\ Cosine Similarity\end{tabular}}} & \multicolumn{1}{c}{\multirow{2}{*}{\name}}                                \\
                                     & \multicolumn{1}{c}{}                                                                     & \multicolumn{1}{c}{}                                                     & \multicolumn{1}{c}{}                                                          & \multicolumn{1}{c}{}      & \multicolumn{1}{c}{}                                                                                           & \multicolumn{1}{c}{}                                                     \\
\rowcolor{intgray} \# of \llm{} Calls at Inference                      & 1                                                                                   & 1                                                                     &      1                                                                      &  1                                                                                                                                                       & N                                                                                               & 1                                                                        \\
\multirow{2}{*}{Input Data}          & \multirow{2}{*}{\begin{tabular}[c]{@{}c@{}}Task-Specific \\ Demonstrations\end{tabular}} & \multirow{2}{*}{\begin{tabular}[c]{@{}c@{}}Single \\ Query\end{tabular}} & \multirow{2}{*}{\begin{tabular}[c]{@{}c@{}}Augmented \\ Dataset\end{tabular}} & \multirow{2}{*}{\begin{tabular}[c]{@{}c@{}}Supervised \\ Data\end{tabular}}            & \multirow{2}{*}{\begin{tabular}[c]{@{}c@{}}Task-Specific \\ Demonstrations\end{tabular}}                       & \multirow{2}{*}{\begin{tabular}[c]{@{}c@{}}Single \\ Query\end{tabular}} \\
                                     &                                                                                          &                                                                          &                                                                               &                                 &                                                        &   
\\ \bottomrule
\end{tabular}
}
\caption{Comparing \name{} with the recent related works for generalization, computation efficiency, and key grounding algorithms. N is the task library size.}
\label{table:compare}
\end{table*}

Recently there has been a surge in research on Augmented Language Model~\citep{mialon2023augmented}, which aims to enable models interface existing ``tools'' for various purposes, such as accessing the latest information~\citep{izacard2022few}, interacting with third-party services~\citep{liang2023taskmatrix}, performing precise calculations~\citep{schick2023toolformer}, or reasoning via code~\citep{cheng2022binding, gao2022pal}. The paradigmatic framework of these tool-augmented LM studies generally comprises two steps: selecting a tool and executing it via a generated API call. Consequently, choosing suitable tools is essential for task success.

The existing works teach language models to select tools using either fine-tuning or in-context learning approaches. For example, Toolformer~\citep{schick2023toolformer} is tailored and limited to a predetermined set of tools observed during pre-training. On the other hand, approaches based on in-context learning~\citep{li2023api, paranjape2023art, chen2023chatcot, sun2023adaplanner, yao2022react} rely on many calls to \llm{} and task-specific demonstrations which diminish their cost efficiency and limits their scalability to a large tool library. To address these limitations, we focus on making the query-tool grounding process more \emph{efficient}, \emph{scalable} and \emph{generalizable}.

In this work, we present \name, \textbf{A}ugment language models with \textbf{G}eneralizable and \textbf{E}fficient tool \textbf{R}esolution, a query-tool grounding algorithm that enables efficient use of tools while also allowing for generalization to both new tasks and large tool libraries. 
The \name{} framework (\autoref{Teaser_Figure}) is comprised of two key modules: (i) Query-Tool Grounding and (ii) Execution. In the \emph{query-tool grounding} module, we compute a grounding score comprised of semantic and pattern based evaluations (introduced in \S\ref{sec:gear:details}). The intuition behind the grounding score is to enable comprehensive query-to-query and answer-to-answer comparisons by leveraging tool description and usage examples, respectively. By considering both question and answer perspectives, the final grounding score provides a comprehensive evaluation of the suitability and compatibility between the given queries and the available tools. Then \name{} passes the selected tool and the given query to the 
\emph{execution} module where a \llm{} is prompted to generate the appropriate API call to obtain the ultimate response from the tool. In general, given $n$ tools in a tool library, \name{} makes $(n+1)$ calls to \slm{}s and only $1$ call to \llm{} (Algorithm \ref{Algorithm}).

Compared to all other in-context learning approaches~\citep{li2023api, paranjape2023art}, \name{} significantly reduces the workload on the \llm{} to do tool grounding, subtask decomposition and API call generation across all tools by assigning query-tool grounding to \slm{}. For instance, compared to ART~\citep{paranjape2023art}, \name{} reduces the calls to \llm{} by directing its intermediate calls to an \slm{} (e.g., \gptNeo) leading to $4\times$ reduction in computational cost (FLOPS), while providing higher accuracy (details in \S\ref{subsec:grounding result}; \autoref{table:tool ratio}).

To the best of our knowledge, there is currently no fine-grained algorithm for query-tool grounding, nor have there been comprehensive empirical experiments to assess tool grounding accuracy across various tool library sizes. Thus, we conduct experiments\footnote{
\href{https://github.com/yining610/GEAR}{Code to reproduce our results is available.}} for \name{} on a variety of different downstream tasks and tool libraries. Our experiments demonstrate that, \name{} improves grounding questions to tools, which leads to stronger downstream performance compared to other few-shot or tool-augmented baselines. 
For example, \name{} leveraging \slm s (e.g., \gptNeo{} with 1.3B parameters) consistently achieves high grounding performance on 12 datasets from 6 NLP tasks, resulting in better downstream accuracy than few-shot prompting and ART~\citep{paranjape2023art}. We also provide evidence of the strong generalizability of \name{} to novel tasks, large tool libraries, and different \slm s.

\section{Related Work}
We divide the notable prior works on tool-augmented models into two groups based on how they modify language models: one uses fine-tuning, while the other uses in-context prompting. We also touch upon works in embodied LM applications.
 
\paragraph{Tool Use via Fine-tuning.} There have been some research efforts focusing on training models to use various language tools~\citep{thoppilan2022lamda, komeili-etal-2022-internet, shuster2022blenderbot, khot2021text, khot2022learning}.

More recently,~\citet{schick2023toolformer} proposes Toolformer which uses a self-supervision manner to train LLMs to use Wikipedia, QA, Calculator, Machine Translation, and Calendar tools.~\citet{parisi2022talm} uses a similar self-supervised approach for teaching models to use tools.~\citet{hao2023toolkengpt} treats tools as special tokens of LLM and learns embeddings for them.~\citet{qiao2023making} proposes a two-stage framework that enables the model to learn through feedback derived from tool execution.~\citet{yang2023gpt4tools} employs instruction tuning to enable LLMs to use multimodal tools. 
Although fine-tuning allows somewhat accurate tool grounding among those observed during training, a key issue with the resulting models is that they cannot utilize new tools without retraining, thus hindering models' generalizability to new tools and tasks.

\paragraph{Tool Use via In-Context Learning.} 
Prior work has used in-context prompting of LLMs utilizes prompts to guide language models generating contextually relevant responses, which is generally more generalizable than fine-tuning. Some notable works here include Chain-of-thought~\citep{wei2022chain}, Zero-shot CoT~\citep{kojima2022large}, among others. These, however, have no access or use external tools. 

ART~\citep{paranjape2023art}, and other concurrent studies~\citep{lu2023chameleon,qian2023creator} support accessing new tools through code or assembling tool sequences to generate the final response. Nonetheless, their way of accessing tools relies on extra task-specific information like demonstrations of how a task needs to be divided or conveyed to existing tools. This restricts their generalizability to new tasks that may necessitate new tools or a different combination of tools. Concurrent work \citep{hsieh2023tool} addresses this issue via documental tool descriptions. However, \name{} complements this work in that, our approach also uses tool outputs for more accurate tool grounding.

Another core issue in all these works is the tool grounding mechanism. \citet{lu2023chameleon, qian2023creator}  rely solely on LLM prompting for tool grounding while ART applies cosine similarity query/tool representations for task grounding. However, little is understood about tradeoffs or limits of these approaches, which we explore in our experiments. To address these, our method extends these works and captures both semantic and pattern relationships (introduced in \S\ref{subsec:semantic} and \S\ref{subsec:pattern}) between query and tools. This allows \name{} to successfully identify and utilize unseen tools for low-resource tasks (novel tasks) without the need for additional task information. \autoref{table:compare} compares \name, CoT, Zero-shot CoT, Toolformer, and ART.

\paragraph{Embodied Language Model in Robotics.} 
Recent research has focused on employing language models for robotic agents planning and their communication with the world~\citep{driess2023palm, zhao2023chat, song2022llm, huang2023grounded,vemprala2023chatgpt}.
This is similar to the setup here involving a language model's interaction with external tools.
~\citet{huang2022inner} and~\citet{lynch2022interactive} leverage various sources of human language and textual feedback to guide robots while solving complex tasks. 
\name{} shares the same underlying idea with SayCan~\citep{saycan2022arxiv} which utilizes binary scores for robotic affordance, while \name{} employs a distinct method that is designed for more general tool and task settings.

\section{\name: Generalizable and Efficient Augmented Tool Resolution}
\label{sec:gear:details}

We start with the formal problem statement. 
We are given an input query $Q$ that we aim to solve. 
In addition, we are provided with a tool library $\mathcal{T} \triangleq \setof{(T_1, d_1, \pi_1), (T_2, d_2,\pi_2), \cdots, (T_n, d_n, \pi_n)}$ 
with $n$ tools. Each tool $T_i$ can receive an API call (e.g., a question or a formula) and respond accordingly, often in the form of natural language. If the provided input is unparsable to the tool, it would return an empty response. Each tool is also supplied with its natural language description ($d_i$) and demonstrations ($\pi_i$) 
showing examples of natural language questions parsed by each tool.

\name{} aims to find the most appropriate tool for solving $Q$. As it can be observed in the Algorithm~\ref{Algorithm}, \name{} iterates over the tools (line 2) and scores each tool $i$ with respect to the given question $Q$ (line 5).
This score is a linear combination of two scores, a \emph{semantic} similarity score $S(., .)$ and a \emph{pattern} similarity score $P(., .)$. Semantic score (defined in \S\ref{subsec:semantic}) provides a measure of semantic alignment between the tool description $d_i$ and the given query $Q$. Pattern similarity score (defined in \S\ref{subsec:pattern}) scores the alignment between the responses obtained from \slm{} and each tool, which provides an indication of how closely the tool's output aligns with a preliminary answer. The algorithm ultimately picks the most appropriate tool based on their scores (line 7) and obtains the final tool response via an API call generated by a \llm{} (line8, line9). 

\newcommand{\sample}{\xleftarrow{\text{sample}}}

\begin{algorithm}[ht]
\caption{\name{} Algorithm}
\label{Algorithm}
\small
\begin{flushleft}
        \textbf{Input:} Query $Q$, Tool library $\mathcal{T}$, Small Language Model (\slm), Large Language Models (\llm)\\
        \textbf{Output:} Grounded tool, and answer to the input question 
\end{flushleft}
\begin{algorithmic}[1]
\State $\hat{a} \sample \text{\slm}(Q)$
\For {$(T_i, d_i, \pi_i)$ in $\mathcal{T}$} 
\State $q_i \sample \text{\slm}(\pi_i +  Q)$ \Comment{\textcolor{darkgreen}{Generate API call}} 
\State $\hat{a}_i \leftarrow T_i(q_i)$ \Comment{\textcolor{darkgreen}{Get the tool's response}}
\State $f_i(Q) \leftarrow \gamma S(Q,d_i) + (1-\gamma)P(\hat{a}, \hat{a}_i)$ \Comment{\textcolor{darkgreen}{Score it}} 
\EndFor
\State $\iota \leftarrow \operatorname*{arg\,max}_{i}f_i(Q)$ \Comment{\textcolor{darkgreen}{Select the best tool}} 
  \State $q_{\iota} \sample \text{\llm}(\pi_{\iota} + Q)$ 
  \Comment{\textcolor{darkgreen}{Generate API call}}
\State $a_\iota \leftarrow T_{\iota}(q_{\iota})$ \Comment{\textcolor{darkgreen}{API call to the selected tool}} 
\State \textbf{Return} grounded tool $T_{\iota}$ and the final answer ${a_\iota}$. 
\end{algorithmic}
\end{algorithm} 

\begin{figure*}[t]
    \centering
    \includegraphics[width=\linewidth]{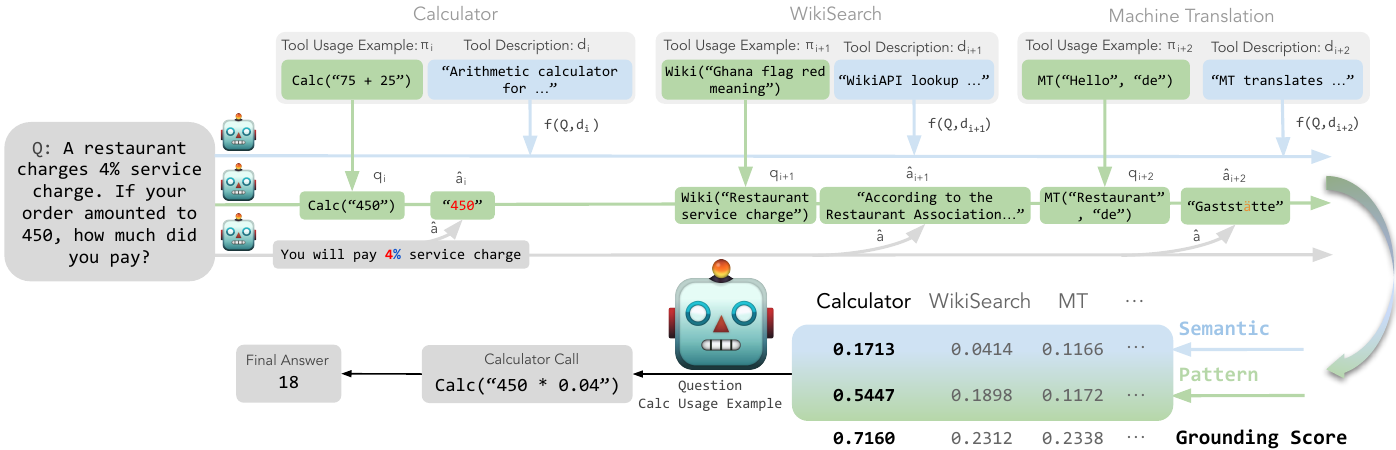}
    \caption{
    \name{} framework. It computes the pattern score by comparing the preliminary answer (in \textcolor{gray}{gray} line) to tool responses (in \textcolor{ForestGreen}{green} box) and the semantic score by comparing the query to tool descriptions (in \textcolor{CornflowerBlue}{blue} box). Grounding tool with the highest weighted average score and executing it via a \llm{} to obtain the final answer.} 
    \label{figure:framework}
\end{figure*}

\subsection{Semantic Similarity Score}
\label{subsec:semantic}
Semantic similarity measures the alignment between the provided question to the language description of a tool. 
For instance, in \autoref{figure:framework}, the description of \calc{} is semantically closer to a query that contains numbers, leading to a higher semantic score. 
Formally, this score is defined as: 
$$
S(Q,d_i) = f_{\text{\slm}}(Q, d_i),
$$
where $f$ is a similarity function utilizing the representation of \slm, quantifying the degree to which the query $Q$ is semantically close to the tool description $d_i$. A popular choice to implement this similarity function (used in our experiments) is cosine distance between the representations query $Q$ and tool description $d_i$:
$$
S(Q,d_i) = \text{cos}\left(\text{enc}_\text{\slm}(Q), \text{enc}_\text{\slm}(d_i) \right),
$$
where $\text{enc}_\text{\slm}(.)$ is the representation of \slm.

\subsection{Pattern Similarity Score}
\label{subsec:pattern}
Pattern similarity provides an answer-level alignment score. This score computes an alignment between a preliminary guess $\hat{a}$ and the response generated by each tool $\hat{a}_i$.  For instance, in \autoref{figure:framework}, the preliminary answer is ``4'', which has a higher pattern similarity score with \calc's response  (``450'', denoted in \textcolor{red}{red}), as both are numbers. Whereas, the responses from \wiki{} and \mt{} are descriptive responses with a large proportion of English tokens (in \textcolor{black}{black}) and a non-ASCII token (in \textcolor{orange}{orange}) that is not exhibited in the preliminary answer. Pattern similarity is computed based on the following steps. 
\paragraph{Preliminary guess.}
First, \slm{} generates a zero-shot preliminary answer $\hat{a}$ for the given query using greedy decoding (line 1).\footnote{We recommend greedy decoding for this zero-shot \slm-based step to reduce the risk of significantly poor responses which may occur in stochastic decoding. 
}

\paragraph{Tool-based response.}
Then \slm{} is prompted by the given query and few shot usage examples to obtain API call $q_i$: 
\begin{align*}
q_i &\sample \text{\slm}(\pi_i + Q).
\end{align*}
We then obtain the tool response $\hat{a}_i \leftarrow T_i(q_i)$ if $q_i$ is parsable by the tool $T_i$, otherwise empty. 

\paragraph{Scoring the alignment.}
The scoring is based on a predefined pattern set $\mathcal{S}$ consisting of distinct elements that correspond to output patterns of various tools. These pattern elements, for example, can represent numbers, English words, symbols, URLs, or certain robotic movements.\footnote{While our evaluation is focused on language tools, the idea discussed here should in principle generalize to other modalities such as physical tools.}
We encode raw tool response $\hat{a}_i$ to its corresponding pattern set $\{e_j(t) \mid \forall j\in \{1,2,\cdots,|\mathcal{S}|\}, \forall t\in \hat{a}_i\}$, where $t$ is the word token of $\hat{a}_i$ and the encoding function $e_j:t\rightarrow \mathcal{S}$ encodes word token to the $j^{th}$ pattern of $\mathcal{S}$ if token exhibits that pattern, otherwise empty.\footnote{For instance, if $\mathcal{S}=\texttt{\{e,f,n\}}$ consisting of English, non-ASCII and number patterns respectively, the sentence ``Hello World 2023'' would be encoded to \texttt{\{e,e,n\}}. If multiple patterns are exhibited in one word token, each pattern would be encoded separately: the German word ``\texttt{lächeln}'' $\Longrightarrow$\texttt{\{e,f,e\}}.} Formally, the output of $e_j$ for $t$ is either a multiset of $j^{th}$ pattern ($\{\mathcal{S}_j^{1},\cdots,\mathcal{S}_j^{n}\}$ where $n\geq 1$) or an empty set $\phi$. Thus, the final encoded pattern set of $\hat{a}_i$ is the multisubset of $\mathcal{S}$. The encoding of $\hat{a}$ follows the same procedure. Let $C_j^{\hat{a}}$ and $C_j^{\hat{a}_i}$ denote the number of $j^{th}$ pattern encoded by $e_j$ in the pattern set of $\hat{a}$ and $\hat{a}_i$. Namely, for $\hat{a}_i$, $C_j^{\hat{a}_i}=|\{e_j(t)\mid \forall t\in \hat{a}_i\}|$. 
Let $|\hat{a}|$ and $|\hat{a}_i|$ be the length of final encoded pattern sets of $\hat{a}$ and $\hat{a}_i$. The pattern similarity score between tool response $\hat{a}_i$ and preliminary answer $\hat{a}$ is computed as:
$$
P(\hat{a}, \hat{a}_i) = \sum_{j\in \{1,\cdots,|\mathcal{S}|\}} \frac{(C_j^{\hat{a}} + \lambda)C_j^{\hat{a}_i}}{(|\hat{a}| + \lambda|\mathcal{S}|)|\hat{a}_i|}\log \frac{1}{\mathcal{P}_j},
$$
where $\mathcal{P}_j$ is the prior probability of the $j^{th}$ pattern from a prior pattern distribution $\mathcal{P}$. $\mathcal{P}, \mathcal{S}$ and $e_j$ can be shared across different task and tool library settings. Add-$\lambda$ smoothing is applied to solve the pattern zero-frequency issue. However, if $\hat{a}_i$ is empty, $P(\hat{a},\hat{a}_i)$ will be assigned its lower bound value 0. In our experiment, we use regular expressions as encoding functions $e_j$.

Intuitively, the pattern similarity score $P(\hat{a},\hat{a}_i)$ is the cross entropy between the prior pattern distribution $\mathcal{P}$ and the smoothed joint pattern distribution from true tool response $\hat{a}_i$ and preliminary answer $\hat{a}$. It is proved to have strict lower and upper bounds in Appendix~\ref{Appendix: Pattern similarity proof} and holds the following five essential properties: (i) \emph{Order Insensitive} (ii) \emph{Length Insensitive} (iii) \emph{Pattern Sensitive} (iv) \emph{Pattern Set Size Insensitive} (v) \emph{Commutative}. Explanations and proofs of these properties are provided in Appendix~\ref{Appendix: Pattern Similarity Score Properties Illustration}. 

We hypothesize that tools could easily elicit their latent pattern distribution through parsable API calls, irrespective of its correctness. Therefore, despite their less reliable performance, \slm{}s are sufficient for query-tool grounding, because their key task is to generate appropriate response patterns in $\hat{a}$ for the given query and parsable API call $q_i$ for the target tool, which is much simpler than reasoning to make $\hat{a}$ (zero-shot result without tool use) or $q_i$ (API call for result with tool use) correct. In Appendix~\ref{Appendix: Mock pattern}, we discuss mock responses which can further enhance the efficiency and generalizability of the grounding process.
\begin{table}[t]
\begin{threeparttable}
\small
\centering
\setlength{\tabcolsep}{1pt}
\begin{tabular}{lcccc}
\toprule
\multirow{4}{*}{\begin{tabular}[c]{@{}l@{}}Algorithm $\rightarrow$ \\ Grounding Model $\rightarrow$ \\Execution Model $\rightarrow$\\Datasets $\downarrow$\end{tabular}} &\multirow{3}{*}{\begin{tabular}[c]{@{}c@{}}Zero-shot\\ $\rule[0.5ex]{0.5em}{0.4pt}$\\ \gptj \end{tabular}} & \multirow{3}{*}{\begin{tabular}[c]{@{}c@{}}Few-shot\\$\rule[0.5ex]{0.5em}{0.4pt}$ \\ 
\gptj \end{tabular}} & \multirow{3}{*}{\begin{tabular}[c]{@{}c@{}} ART$^\ast_\text{llm}$ \\ \gptNeo \\ \gptj \end{tabular}} & \multirow{3}{*}{\begin{tabular}[c]{@{}c@{}} \name\\ \gptNeo{}\\ \gptj \end{tabular}} \\
                          &&&&\\
                          &&&&\\
                          &&&&\\
                         \midrule
ASDiv      & 7.5        & 21.4          & 16.7                & 23.3              \\
GSM8K        & 0.4      & 5.6           & 9.8                   & 3.8           \\
SVAMP       & 2.0        & 13.1          & 11.2                 & 18.6           \\ \hdashline\noalign{\vskip 0.5ex}
\arrowdownright \footnotesize \textbf{Average (Arithm)}                 & 3.3           & 13.4          & 12.6                 & \textbf{15.2}                \\ 
\midrule
IWSLT (cn)     & 10.5    & 16.9          & 4.1                   & 21.1                 \\
IWSLT (ar)      & 8.5      & 18.7          & 4.8                  & 17.6                 \\
IWSLT (de)     & 7.7       & 19.3          & 5.4                   & 32.9                \\
IWSLT (fr)    & 7.9      & 22.7          & 6.7                   & 38.4                 \\
IWSLT (ja)    & 5.5       & 14.4          & 3.4                   & 12.9                 \\
IWSLT(ko)    & 8.9        & 15.2          & 3.6                   & 14.9                \\ \hdashline\noalign{\vskip 0.5ex}
\arrowdownright \footnotesize \textbf{Average (MT)}       & 8.2                           & 17.9          & 4.7                 & \textbf{23.0}          \\ 
\midrule
NQ-Open     & 10.2          & 31.1          & 21.2                & 43.4                \\
WebQS    & 5.3            & 18.2          & 11.2               & 22.1                   \\
TriviaQA     & 27.3         & 46.5          & 29.3                 & 50.3               \\
\hdashline\noalign{\vskip 0.5ex}
\arrowdownright \footnotesize \textbf{Average (ODQA)} &14.3                           & 31.9          & 20.6               & \textbf{38.6}             \\ 
\midrule
CSQA       & 10.9          & 37.1          & 6.3                 & 60.7             \\
COPA        & 6.5
           & 27.0          & 1.0          & 13.6       \\
SocialIQA    & 8.4
          & 26.0          & 5.5                 & 41.5  \\ 
          \hdashline\noalign{\vskip 0.5ex}
\arrowdownright \footnotesize \textbf{Average (CSQA)}            & 8.6
  & 30.0          & 4.3               & \textbf{38.6}          \\ \bottomrule
\end{tabular}
\caption{Downstream task performance results (\S\ref{subsec:downstream}). Evidently, \textbf{\name{}-augmented \gptj{} outperforms our baselines} when using a consistent set of grounding and execution models. 
}
\label{table:acc}
\end{threeparttable}
\end{table}

\begin{table*}[t]
\small
\centering
\begin{tabular}{lcccccc}
\toprule
        \textbf{Models}              & \textbf{ASDiv} & \textbf{SVAMP} & \textbf{SQuAD} & \textbf{T-REX} & \textbf{TriviaQA} & \textbf{MLQA(es)} \\ \midrule   
\textbf{Toolformer (\gptj)} & 40.4           & 29.4           & 33.8           & 53.5           & 48.8              & 20.6              \\
\textbf{ART$^\ast_{\text{llm}}$ (\gptNeo/\gptThree)} & 37.0 & 21.3 & 17.7 & 20.6 & 24.3 & 14.0 \\
\textbf{ART$_{\text{cs}}$ (\miniLM/\gptDavinciThree)}      & \textbf{86.7}           & 77.3           & 39.3           & 50.4           & 61.0              &      $\rule[0.5ex]{0.5em}{0.4pt}$              \\
\textbf{\name{} (\gptNeo/\gptDavinciThree)}     & 74.9 (-11.8)           & \textbf{79.9 (+2.6)}           & \textbf{61.1 (+21.8)}           & \textbf{83.1 (+32.7)}           & \textbf{62.5 (+1.5)}              & \textbf{58.3 (+37.7)}              \\ \bottomrule
\end{tabular}
\caption{Comparing \name{} with Toolformer~\citep{schick2023toolformer} and ART~\citep{paranjape2023art} (\S\ref{subsec:downstream}). The original ART work, ART$_{\text{cs}}$, employs \miniLM{} for cosine similarity strategy and does not have \qa{} or \mt{} for the MLQA task.}
\label{table:Toolformer comparison}
\end{table*}
\begin{table*}[t]
\small
\centering
\begin{tabular}{cccccccc}
\toprule
\multicolumn{1}{l}{\textbf{Models}}                      & \textbf{\begin{tabular}[c]{@{}c@{}}Evaluate on $\rightarrow$\\ Demonstration $\downarrow$\end{tabular}} & ASDiv                    &GSM8K                    &SVAMP                    & TriviaQA                 & NQ-Open                  & WebQS                    \\ \midrule
\multicolumn{1}{c}{\multirow{6}{*}{\textbf{ART$_{\text{cs}}$ (\miniLM/\gptDavinciThree)}}} & \multicolumn{1}{c}{ASDiv}                                                                & \cellcolor{lightblue}97.9                     & \cellcolor{lightblue}88.5                     & \cellcolor{lightblue}87.2                     & \cellcolor{lightred}2.1                      & \cellcolor{lightred}1.4                      & \cellcolor{lightred}0.0               \Tstrut      \\
\multicolumn{1}{c}{}                     & \multicolumn{1}{c}{GSM8K}                                                               & \cellcolor{lightblue}93.8                     & \cellcolor{lightblue}88.4                     & \cellcolor{lightblue}81.9                     & \cellcolor{lightred}0.3                      & \cellcolor{lightred}1.1                      & \cellcolor{lightred}0.0                    \\
\multicolumn{1}{c}{}                     & \multicolumn{1}{c}{SVAMP}                                                               & \cellcolor{lightblue}98.3                     & \cellcolor{lightblue}74.5                     & \cellcolor{lightblue}75.7                     & \cellcolor{lightred}0.0                       & \cellcolor{lightred}1.1                      & \cellcolor{lightred}0.0                     \\
\multicolumn{1}{c}{}                     & \multicolumn{1}{c}{TriviaQA}                                                             & \cellcolor{lightred}25.8                     & \cellcolor{lightred}32.2                     & \cellcolor{lightred}22.5                     & \cellcolor{lightblue}98.1                     & \cellcolor{lightblue}96.2                     & \cellcolor{lightblue}0.4                    \\
\multicolumn{1}{c}{}                     & \multicolumn{1}{c}{NQ-Open}                                                             & \cellcolor{lightred}25.3                     &       \cellcolor{lightred}25.2                   & \cellcolor{lightred}22.4                     & \cellcolor{lightblue}97.4                     & \cellcolor{lightblue}98.2                     & \cellcolor{lightblue}0.4                 \\
\multicolumn{1}{c}{}                     & \multicolumn{1}{c}{WebQS}                                                                &    \cellcolor{lightred}28.6                      & \cellcolor{lightred}39.9                     & \cellcolor{lightred}28.3                     & \cellcolor{lightblue}94.8                     & \cellcolor{lightblue}96.8                     & \cellcolor{lightblue}1.1                     \Bstrut  \\ \midrule
\multicolumn{1}{c}{\textbf{\name{} (\gptNeo/\gptDavinciThree)} }                   & \multicolumn{1}{c}{}    & 83.1 & 83.0 & 89.0 & 63.0 & 65.6 & 54.3 \\ \bottomrule
\end{tabular}
\caption{Cross-dataset generalization evaluation of tool grounding accuracy (\S\ref{subsec:grounding result}). Evidently,  \textbf{\name{} can identify the appropriate tool for a given task without requiring \colorbox{lightblue}{in-domain} demonstrations} while ART has a significant grounding performance decline on \colorbox{lightred}{out-domain} demonstrations, with each score representing grounding accuracy/affordance ratio in percentage.}
\label{table:generalization test}
\end{table*}
\section{Experiment Setup}
\label{section:experiment:setup}
\subsection{\name{} Implementation.}

We implement \name{} according to the construction described in \S\ref{sec:gear:details}. 
Throughout the experiments the \llm s  in our study are \gptj{} and \gptDavinciThree{} (in short, \gptThree{}), and our \slm s are \gptNeo{}, \gptTwo{}, \gptTwoLarge{}, \miniLM{} and \mpnet.\footnote{We accessed the OpenAI models on April through June, 2023.}

Specifically for our implementation of \name, we use \mpnet{} to calculate semantic similarity scores and \gptNeo{} for generating preliminary answers and API calls to calculate pattern similarity scores. For \llm s, we use either \gptj{} or \gptThree{} for final tool execution. 


\paragraph{Tools.} To evaluate the performance for a variety of purposes, we create a total of 10 different tools, including 4 basic tools: \calc{}, \mt{}, \wiki{}, and \qa{}; and 6 novel tools: \tz{}, \mlqa{}, \sleep{}, \expcalc{}, \logcalc{}, and \map{}. All of them are accessible via specific API calls and have corresponding returns. Examples of API calls are shown in \autoref{table:apicall} and more information about tools can be found in Appendix~\ref{Appendix:tools}.

\paragraph{Datasets.} We conduct our experiment on 14 datasets across 6 downstream tasks. The dataset details and evaluation metrics can be found in Appendix \ref{Appendix:datasets}.    

\subsection{Baseline Systems}
\label{subsec:systems}
We organize our baselines as follows:
\begin{itemize}[leftmargin=0.1in]
    \item \textbf{Zero-shot}: This baseline directly asks questions to \llm{} without any instruction.
    \item \textbf{Few-shot}: This baseline involves prompting \llm{} with natural language instructions that articulate the requirements of the given task.
    \item \textbf{ART}: This approach uses prompting \llm{} for multi-step reasoning and tools execution~\cite{paranjape2023art}. Besides the results in the original paper, we experiment with a reimplementation of ART (referred to as ART$^\ast$) adapted to our tools and tasks. 
    Specifically, following the original work, we report two variants of this model with different tool-grounding strategies proposed in its paper:  
    (1) \llm{}-based prompting similarity (ART$^\ast_{\text{llm}}$) 
    and (2) cosine similarity (ART$^\ast_{\text{cs}}$).
\end{itemize}
To ensure a fair comparison between baselines, we let few-shot, ART$^\ast$, and \name{} use the same prompt examples (Appendix~\ref{Appendix:prompts}).

\section{Experimental Findings}
\label{section:experiment}
We compare the downstream performances of models (\S\ref{subsec:downstream}), and compare their generalizability to new tools or tasks (\S\ref{subsec:grounding result}).
\begin{table*}[t]
\small
\centering
\begin{tabular*}{\textwidth}{@{\extracolsep{\fill}} lccccccc}
\toprule
\textbf{Algorithm} $\rightarrow$ &  & \multicolumn{3}{c}{\textbf{\name}}                  &  ART$^\ast_{\text{llm}}$ & ART$^\ast_{\text{llm}}$ & ART$^\ast_{\text{cs}}$           \\ \cmidrule(r){3-5} \cmidrule(r){6-8} \multirow{2}{*}{\begin{tabular}[c]{@{}l@{}}\textbf{Grounding Model} $\rightarrow$\\ \textbf{Dataset (w/ 4 Tools)} $\downarrow$\end{tabular}} & \multirow{2}{*}{\begin{tabular}[c]{@{}l@{}} \\ \textbf{Target Tool} $\downarrow$\end{tabular}} & \multirow{2}{*}{\begin{tabular}[c]{@{}c@{}}\gptNeo\\(1.3B)\end{tabular}} & \multirow{2}{*}{\begin{tabular}[c]{@{}c@{}}\gptTwoLarge\\(774M) \end{tabular}}& \multirow{2}{*}{\begin{tabular}[c]{@{}c@{}}\gptTwo\\(355M)\end{tabular}} & \multirow{2}{*}{\begin{tabular}[c]{@{}c@{}}\gptNeo\\(1.3B)\end{tabular}} & \multirow{2}{*}{\begin{tabular}[c]{@{}c@{}}\gptDavinciThree\\(175B)\end{tabular}} & \multirow{2}{*}{\begin{tabular}[c]{@{}c@{}}\mpnet\\(110M)\end{tabular}}\\
&&&&&& \\\midrule
ASDiv            & \tt{Cal}                   & 83.1       & 77.7      & 58.7      & 25.6       & 46.5 & 98.8 \\
GSM8K            & \tt{Cal}                       & 83.0      & 65.3      & 55.6      & 38.0       & 45.5 & 99.5 \\
SVAMP            & \tt{Cal}                    & 89.0        & 76.5      & 65.1      & 21.0      & 50.0 & 100.0 \\ \hdashline\noalign{\vskip 0.5ex}
\arrowdownright \textbf{Average (Arithm)}         &                      &\textbf{85.0}       & 73.2     & 59.8      & 28.2  & 47.3 & \underline{\textbf{99.4}}       \\  \midrule
IWSLT (cn)      & \tt{MT}                           & 84.1        & 95.5      & 98.2      & 30.0        & 63.2 & 99.9 \\
IWSLT (ar)      & \tt{MT}                           & 66.6        &   $\rule[0.5ex]{0.5em}{0.4pt}$          &      $\rule[0.5ex]{0.5em}{0.4pt}$       & 27.8       & 61.6 &  98.6\\
IWSLT (de)      & \tt{MT}                            & 96.9        & 94.4      & 95.2      & 31.6     & 66.0 & 94.0  \\
IWSLT (fr)     & \tt{MT}                             & 96.6        & 94.0      & 96.0      & 33.8     & 64.4 &  92.2 \\
IWSLT (ja)    & \tt{MT}                              & 72.4        & 89.3      & 91.1      & 30.8     & 62.8 & 97.8  \\
IWSLT (ko)   & \tt{MT}                               & 82.2        & 66.7      & 91.7      &  25.9      & 72.7 &   99.4    \\ \hdashline\noalign{\vskip 0.5ex}
\arrowdownright \textbf{Average (MT)}    &                              & 83.1        & 88.0      & \textbf{94.4}      & 30.0    & 65.1  & \underline{\textbf{97.0}}    \\ \midrule
NQ-Open      & \tt{Wiki}                        & 63.0        & 61.3      & 59.1      & 10.9      & 44.0 & 39.4 \\
WebQS        & \tt{Wiki}                       & 65.6        & 83.1      & 81.4      & 13.6     & 56.8 &  60.5 \\
TriviaQA     & \tt{Wiki}                        & 54.3        & 77.2      & 71.7      & 13.2     & 58.1 &  41.5 \\ \hdashline\noalign{\vskip 0.5ex}
\arrowdownright \textbf{Average (ODQA)}    &                             & 61.0        & \underline{\textbf{73.9}}      & 70.7      & 12.6  & \textbf{53.0}   & 47.1    \\ \midrule
CommonsenseQA   & \tt{QA}                             & 77.1        & 84.0      & 84.9      & 10.1     & 34.9 & 69.7  \\
COPA       & \tt{QA}                                   & 41.3        & 77.2      & 61.2      & 7.2     & 24.4 &  29.7  \\
SocialIQA      & \tt{QA}                               & 75.7        & 87.6      & 59.5      & 16.4      & 42.4 & 14.1 \\ \hdashline\noalign{\vskip 0.5ex}
\arrowdownright \textbf{Average (CSQA)}  &                     & 64.7        & \underline{\textbf{82.9}}      & 68.5      & 11.2  & \textbf{33.9} & 37.8  \\ \midrule
\multicolumn{2}{l}{\textbf{\# of Operation in GFLOPS$^6$}}  & 1573 & 937 & 430 & 5455 & 728420\footnotemark& 160 \\
\bottomrule
\end{tabular*}
\vspace{-0.0cm}
\caption{Tool grounding accuracy for 4 downstream tasks with a 4-tools library (\S\ref{subsec:grounding result}). \textbf{Bold} denotes the highest value within its grounding strategy and \underline{\textbf{underline}} represents the highest among all baselines. We find that \textbf{\name{} yields better performance compared to the \llm{}-based strategy on all datasets. \name{} is generalizable to smaller \slm{}s and even achieve better grounding results on certain tasks}.}
\label{table:tool ratio}
\end{table*}
\addtocounter{footnote}{0}

\subsection{Results on Downstream Tasks}
\label{subsec:downstream}
We first evaluate all our models on the downstream task performance with a tool library containing 4 basic tools (\autoref{table:acc}). For consistency of comparisons, all the baselines use \gptj{} for the final answer execution. \name{} outperforms all the baselines across four basic tasks. For example, the accuracy of \name{}-augmented \gptj{} is $24.3\%$ and $6.7\%$ higher than zero-shot and few-shot baselines on the ODQA (Open-domain QA) task. Compared to the ART$^\ast_{\text{llm}}$, \name{} consistently has superior performance because of better tool use. Later in \S\ref{subsec:grounding result} we show that this performance gap is due to the difference in tool grounding accuracy. Additional results using \gptThree{} as execution model (in place of \gptj) are provided in Appendix~\ref{Appendix: Performance Experiment}.

\autoref{table:Toolformer comparison} puts Toolformer~\citep{schick2023toolformer}, ART~\citep{paranjape2023art} and \name{} together, evaluating on their shared datasets. All datasets are evaluated under a 4 basic tools library except for MLQA which uses a 5-tools library with an extra \mlqa{} tool.
Since Toolformer code and model are not available online, we are not able to reproduce their results and therefore, copy the numbers from its paper. The comparison is unfair to Toolformer as it uses a finetuned \gptj{} model. 
But it is informative that \name{}-augmented \gptThree{} outperforms the original work ART$_{\text{cs}}$, which employs the same-sized model with task-specific demonstrations, on 4 out of 5 tasks. This performance gain also emphasizes the strong generalization capability of \name{}.

\subsection{Results on Tool Grounding}
\label{subsec:grounding result}
\begin{figure}[t]
    \centering
    \includegraphics[width=0.48\textwidth,trim=0.3cm 0.8cm  -.1cm 0.2cm]{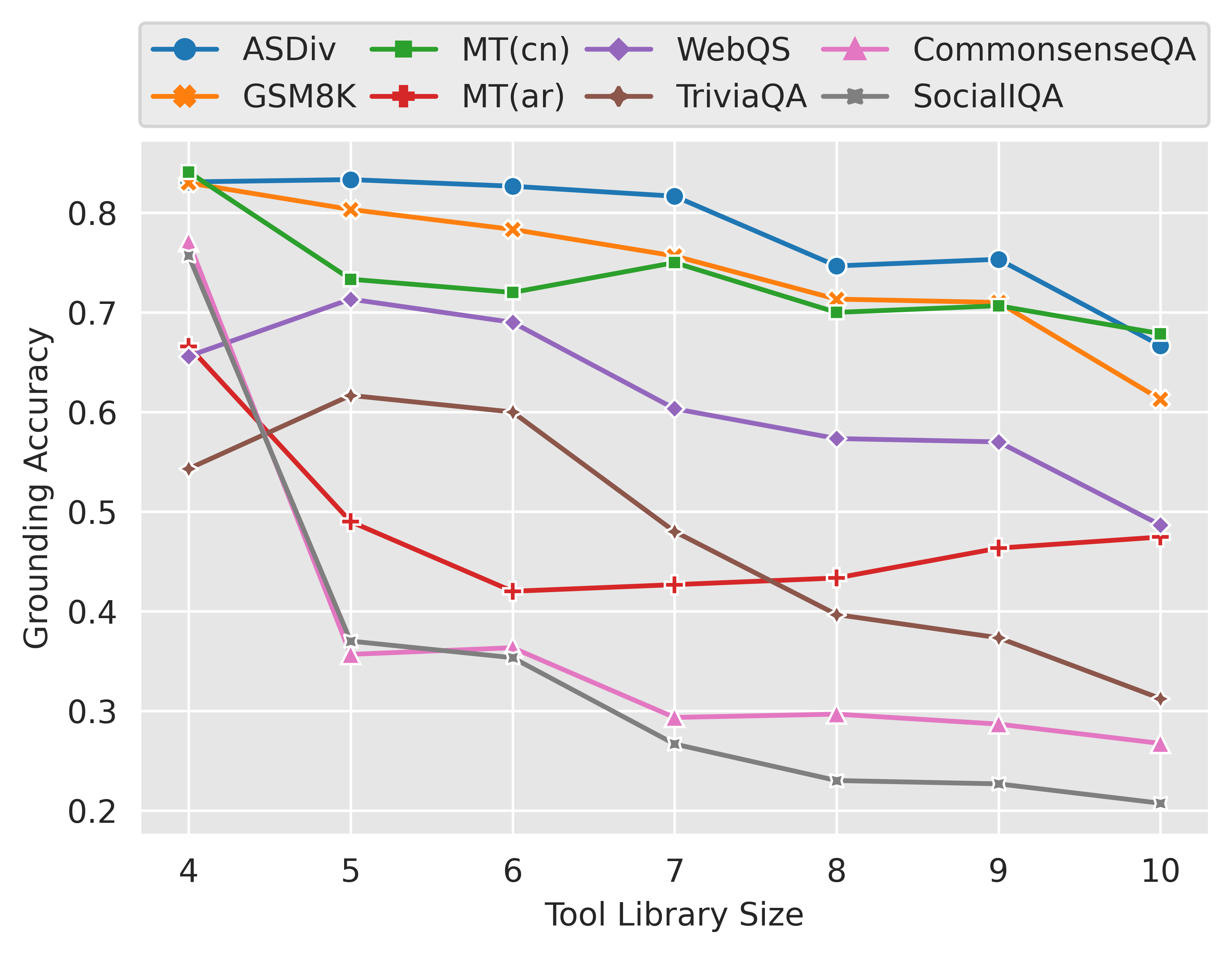}
    \caption{
    Grounding accuracy of \name{} 
    when the tool library is expanded from 4 to 10 tools (\S\ref{subsec:grounding result}). We incrementally incorporate these tools: \mlqa{}, \tz{}, \sleep{}, \logcalc{}, and \map{}.}
    \label{fig:grounding}
\end{figure}

\footnotetext{Since OpenAI has not open sourced their \gptDavinciThree, we approximate the operations as $\text{\# tokens} \times \text{\# params}$, which is the lower bound of operations. The real amount of operations should exceed this estimation.}
We systematically examine the tool grounding accuracy (the percentage of correctly selected tools) across a variety of tool library sizes and model sizes. We first calculate the grounding accuracy for a tool library comprising 4 basic tools. Then we expand the tool library to a total of 10, as described in Appendix~\ref{Appendix: novel tools}, by introducing competitor and distractor tools. We re-evaluate the grounding accuracy for the four basic tasks, along with two novel tasks requiring \mlqa{} and \tz{} tools. The main results are shown in \autoref{table:tool ratio} and \autoref{fig:grounding}. 

\paragraph{\name{} is more generalizable than other query-tool grounding algorithms.} According to \autoref{table:tool ratio}, \name{} utilizing \gptNeo{} with 1.3B parameters significantly outperforms the \llm{}-based strategy proposed by ART~\citep{paranjape2023art}, even when the latter uses \gptThree{} which is 134 $\times$ larger. The best-reported similarity strategy in ART, which calculates the cosine similarity between the given demonstration and textual description of tasks, performs outstandingly well on Arithmetic and MT tasks. We hypothesize this is because of the presence of distinct and unique keywords in Arithmetic and MT queries, which are easily distinguishable by word embeddings. However, for more open-ended NLP tasks like Open-domain and Commonsense QA, word embeddings are less generalizable in selecting the correct tools, resulting in low grounding accuracy of $47.1\%$ and $37.8\%$. In contrast, \name's grounding strategy is shown to be more strong with grounding accuracy of $61.0\%$ and $64.7\%$ on the aforementioned tasks.

\autoref{table:generalization test} displays a substantial decline in grounding accuracy of ART~\citep{paranjape2023art} when using out-domain demonstrations. In contrast, \name{} consistently maintains its high performance without requiring in-domain demonstrations. We also demonstrate \name{} outperforms retrieval-based baselines on query-tool grounding, as shown in \autoref{table:retrieval baselines} in Appendix \ref{Appendix:grounding experiment}.

\paragraph{\name{} is generalizable to smaller language models.} We evaluate the grounding performance of \name{} on two smaller GPT-2 models. As reported in \autoref{table:tool ratio}, \name{} consistently exhibits high-level grounding accuracy on both \slm{}s and even outperforms \gptNeo{} on certain tasks. For example, \name{}-augmented \gptTwoLarge{} achieves $73.9\%$ and $82.9\%$ grounding accuracy for the Open-domain QA and Commensense QA tasks, greatly higher than those of ART$^\ast$ baselines. Moreover, as the model size increases, the marginal grounding accuracy gain diminishes. This is because as long as the \slm{} produces expected patterns for the given query, the correctness of the preliminary answer has no bearing on the pattern similarity score (see case study in \S\ref{Appendix: case study}). Which, in turn, experimentally proves the feasibility of employing \slm{}s for query-tool grounding. 
\paragraph{\name{} is generalizable to larger tool libraries.} Because of a more comprehensive grounding process, \name{} enables certain tasks to generalize better for larger sets of tools.
\autoref{fig:grounding} displays the grounding accuracy changing from 4 to 10 tools. The general low decreasing rates for Arithmetic, MT and Open-domain QA demonstrate the ability of \name{} in handling tool libraries of varying sizes.

We hypothesize the drops between the fourth and fifth tools of CommonsenseQA and SoicalIQA datasets are likely due to the introduction of the \mlqa{} tool which has functional overlap with the basic \qa{} tool. Specifically, the \mlqa{} tool can also solve reasoning tasks by translating contexts from English to English; therefore, if we consider \mlqa{} as the correct tool for the Commonsense QA task as well, the averaged final grounding accuracy of Commonsense QA task will increase to $49.1\%$, with a $15.6\%$ decrease compared to~\autoref{table:tool ratio}.

We also compare \name{} and the best variant ART$^\ast_{\text{cs}}$ under a 10-tools library on 6 downstream tasks with two extra novel tasks. In short, \name{} outperforms ART$^\ast_{\text{cs}}$ on 5 out of 6 tasks. See Appendix~\ref{Appendix:grounding experiment} for detailed results.

\begin{table}[t]
\small
\centering
\begin{tabular*}{\linewidth}{@{\extracolsep{\fill}}lccc}
\toprule
\multirow{2}{*}{\textbf{Task}} & \multirow{2}{*}{\textbf{\name{}}}  & \multicolumn{2}{c}{Performance change $\Delta$}  \\ 
     &      & $\neg$Pattern Sim   & $\neg$Semantic Sim  \\ \midrule
Arithm            & 74.0          & \textbf{-2.3}	&\textbf{-11.5}
 \\
MT        & 80.5      & +10.9 &	-69.9
 \\
ODQA      & 40.7              & \textbf{-15.4}& \textbf{-21.1}
  \\
CSQA      & 33.4             & \textbf{-21.6}	& \textbf{-18.9} \\
MLQA      & 54.4            & \textbf{-10.6}	& \textbf{-31.5} \\
TZ Conversion   & 96.4           & +3.6	& -94.9
 \\ \bottomrule
\end{tabular*}
\caption{The result of leave-one-out ablation study for 10-tools library (\S\ref{subsec:ablation study}). \textbf{The decrease in grounding accuracy on both columns demonstrates the importance of considering both semantic and pattern scores for query-tool grounding.}}
\label{table:ablation study 10 tools}
\end{table}


\section{Analysis}
\subsection{Ablation Study}
\label{subsec:ablation study}
We now perform a leave-one-out experiment to better understand the contribution of each score (\S\ref{subsec:semantic} and \S\ref{subsec:pattern}) to the final grounding accuracy. We conduct experiments for a 10-tools library with only either semantic similarity score or pattern similarity score. The results are shown in \autoref{table:ablation study 10 tools}. For the 10-tools library, there are 4 out of 6 tasks displaying grounding accuracy decline in both semantic and pattern columns, suggesting that it is crucial to consider both semantic and pattern similarity scores for query-tool grounding. Tasks such as MT and Timezone Conversion show increased grounding accuracy in the semantic column, which is likely due to the same reason discussed in \S\ref{subsec:grounding result}: these two tasks contain unique keywords so that single semantic similarity score suffices to distinguish them from other tasks (more results in Appendix~\ref{Appendix: Ablation Study}.)
\begin{figure*}[t]
\begin{mdframed}[leftmargin=2em, align=center]
\footnotesize
[\textit{Question}] Janet’s ducks lay 16 eggs per day. She eats three for breakfast every morning and bakes muffins for her friends every day with four. She sells the remainder at the farmers' market daily for $\$$2 per fresh duck egg. How much in dollars does she make every day at the farmers' market? \\
\noindent [\textit{Preliminary answer from} \gptTwo{}] i make \colorbox{lightpurple}{\$}\colorbox{lightred}{2}\\
\noindent [\textit{Preliminary answer from} \gptTwoLarge{}] the average american spends \colorbox{lightpurple}{\$}\colorbox{lightred}{1000} a year on food \\
\noindent [\textit{Preliminary answer from} \gptNeo{}] janet’s ducks lay \colorbox{lightred}{16} eggs per day \\
\noindent [\textit{Preliminary answer from} \gptThree{}] Janet eats \colorbox{lightred}{3} of the \colorbox{lightred}{16} eggs, so she has \colorbox{lightred}{16 - 3 = <<16-3=13>>13} eggs left. She bakes muffins with \colorbox{lightred}{4} of these eggs, so she has \colorbox{lightred}{13 - 4 = <<13-4=9>>9} eggs left to sell. Thus, she makes \colorbox{lightpurple}{\$}\colorbox{lightred}{9 * 2} = \colorbox{lightpurple}{\$}\colorbox{lightred}{<<9*2=18>>18} every day at the farmers' market. Answer: \colorbox{lightpurple}{\$}\colorbox{lightred}{18}.
\end{mdframed}
\vspace{-.2cm}
\begin{mdframed}[leftmargin=2em, align=center]
\footnotesize
[\textit{Question}] In which decade did Billboard magazine first publish and American hit chart? \\
\noindent [\textit{Preliminary answer from} \gptTwo{}] the first billboard magazine was published in the fall of \colorbox{lightred}{1892}.\\
\noindent [\textit{Preliminary answer from} \gptTwoLarge{}] the first billboard magazine was published in the united states in the year \colorbox{lightred}{900}. \\
\noindent [\textit{Preliminary answer from} \gptNeo{}] the billboard hot \colorbox{lightred}{100} is a weekly chart that ranks the most popular songs in the United States \\
\noindent [\textit{Preliminary answer from} \gptThree{}] The \colorbox{lightred}{1930}s. Specifically, Billboard magazine first published its "Music Popularity Chart" in \colorbox{lightred}{1936}.
\end{mdframed}
\vspace{-.2cm}
\caption{A comparison of output patterns between \slm{}s and \llm{}. The lines subsequent to [\textit{Question}] represents the output generated by the corresponding model, with patterns (\colorbox{lightred}{number}, \colorbox{lightpurple}{symbol} and \colorbox{white}{English alphabet}) labeled in different colors. While \slm{}s tend to be less accurate than \llm{}, \textbf{their responses provide sufficient clues (pattern distribution) about the form of the expected answer}.}
\label{fig: case study}
\end{figure*}
\begin{figure}[t]
    \centering
    \includegraphics[width=0.48\textwidth,trim=0cm 0.4cm 0cm 0cm]{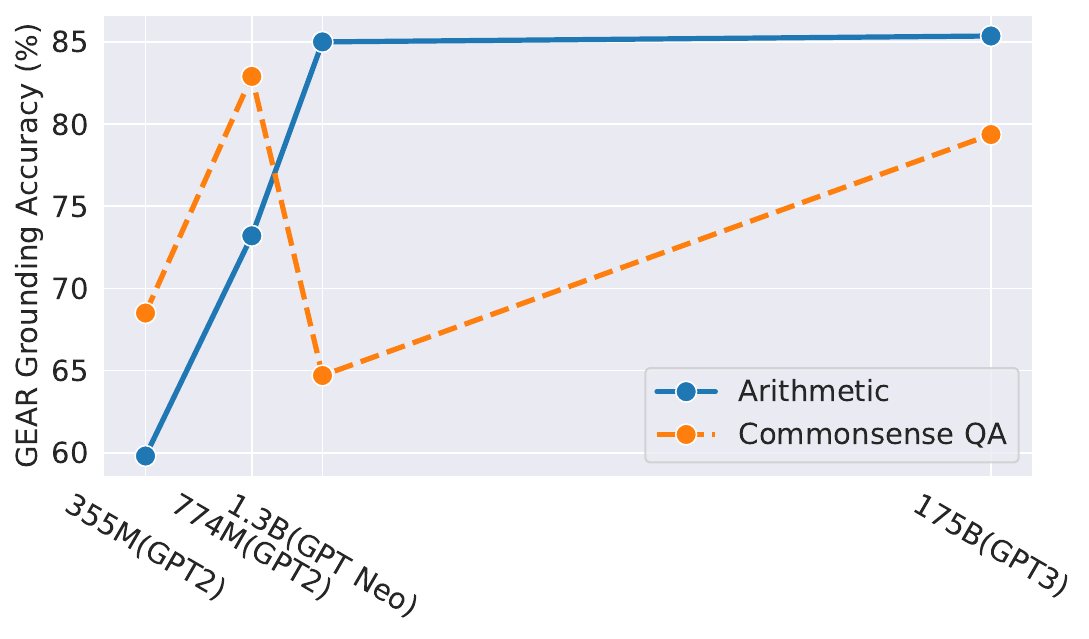}
    \caption{Averaged \name{} grounding performance over \slm{} sizes (number of parameters, in log scale) on Arithmetic and Commonsense QA tasks. Each task is evaluated by three datasets. \textbf{\name{} with \slm{} has a similar grounding accuracy as with \llm{}}.}
    \label{fig:gear grounding performance}
\end{figure}
\subsection{Case Study on \slm's Size}
\label{Appendix: case study}
It is natural to question whether \name{} will have much better performance if we replace \slm{} with \llm{}, namely, $\hat{a} \sample \text{\llm}(Q)$ in Algorithm \ref{Algorithm}. 
We provide a case study (\autoref{fig: case study}) showing the impact of various \slm{} choices, including the setting where replacing \slm{} with \llm{}, to further illustrate our observation in \S\ref{subsec:grounding result} that as the model size increases, the marginal grounding accuracy gain diminishes (\autoref{fig:gear grounding performance}).
In the first example from GSM8K~\citep{cobbe2021gsm8k}, we can see that \slm{} offers the similar indicative signal as \llm{} that the potential answer should contain \colorbox{lightred}{number} and \colorbox{lightpurple}{symbol} patterns, despite their responses being incorrect.  We also observe that this phenomenon not only happens in pattern-specific tasks (e.g. Arithmetic) but also occurs in more general open-ended tasks like Commonsense QA. The second TriviaQA~\citep{joshi-etal-2017-triviaqa} example shows that the pattern distributions generated by the \slm{}s closely resemble the \llm{}'s distribution: a single number amid English text.

Thus as long as API calls are properly generated, it is highly likely that \name{} with \slm{} will select the same tool as with \llm{}. In other words, generating executable API calls from \slm{} now becomes the only empirical limitation of the upper bound of the pattern similarity score. As the model size increases, this limitation will become less strict, resulting in a diminished rate of improvement in grounding performance.

To validate the above observations, we visualize the grounding performance of \name{} across different \slm{} sizes on these two tasks in \autoref{fig:gear grounding performance}. Evidently, as the increasing of \slm{} sizes, the grounding performance margin tends to decrease. Note that because of different model families, \slm{} grounding performance may not necessarily be monotonically increasing (orange line).

\section{Conclusion}
In this paper, we introduce \name: a generalizable query-tool grounding algorithm that enables efficient tool groundings without extra fine-tuning or task-specific demonstrations. 
This is accomplished by introducing a fine-grained scoring mechanism that leverages both semantic and pattern similarities and leveraging smaller language models for query-tool grounding. 
To validate the generalizability of \name{}, we conduct extensive experiments that demonstrate its capacity to deal with large tool libraries and novel tasks.

\section*{Limitations}
While \name{} aims to improve the query-tool grounding and exhibits strong generalization and robustness for large tool libraries, including user-provided pipelines, it has a potential limitation in lacking support for automatic tool pipeline construction. Future works could focus on how to combine \name{} with automatic reasoning and task decomposition works, such as ART~\citep{paranjape2023art}, Chameleon~\cite{lu2023chameleon}, and CREATOR~\citep{qian2023creator}. We believe that the combination of generalizable and efficient tool grounding with multi-hop reasoning would further boost the performance of the current SOTA \llm{}s.  

Theoretically, \name{} supports tools that have non-textual returns via mock responses. However, we only test the \sleep{} and \map{} tools in the main experiment and the \image{} tool in the \name{}-augmented chatbot. Though achieving promising results on these three tools, future works, especially in the embodied LM area, could further explore how mock responses can be used in grounding human language with physical world tools.

\section*{Acknowledgements}
The authors would like to thank Adam Byerly, Tianjian Li, and Zhengping Jiang for their helpful input on earlier versions of this work. 
The authors would like to acknowledge the support of ONR grant N00014-24-1-2089 and the gifts from Amazon and the Allen Institute for AI. 
Moreover, the authors would like to thank the Center for Language and Speech Processing members at Johns Hopkins University for their valuable feedback and comments.
GPU machines for conducting experiments were provided by ARCH Rockfish cluster (\url{https://www.arch.jhu.edu}).

\bibliography{anthology,custom}
\bibliographystyle{acl_natbib}

\onecolumn

\clearpage
\appendix
\begin{center}
{\LARGE \textbf{Supplemental Material}}
\end{center}

\begin{center}
\begin{tabular}{@{}ll@{}}
\toprule
Appendix & Contents \\ \midrule
\autoref{Appendix:Pattern similarity}       & Further illustrations for pattern similarity score: examples and proofs        \\
\autoref{Appendix:implementation details}        & Extra experiment details: hyperparameters, patterns, models and datasets        \\
\autoref{Appendix:tools}       & Tool introduction       \\
\autoref{Appendix: Performance Experiment}        &  Extra results on downstream experiment         \\
\autoref{Appendix:grounding experiment}        & Extra results on tool grounding experiment        \\
\autoref{Appendix: Ablation Study}        &   Extra results on ablation study       \\ 
\autoref{Appendix:chatbot}       &  \name{}-augmented chatbot: implementation details and survey results        \\
\autoref{Appendix:prompts}        &  Prompts used for \name{} and few-shot baseline      \\ \midrule
\end{tabular}
\end{center}

\section{Pattern Similarity Score}
\label{Appendix:Pattern similarity}
\subsection{Pattern Similarity Score Bounds}
\label{Appendix: Pattern similarity proof}
Because the count $C$ and $\lambda$ are nonnegative, $\mathcal{P}_j \in [0,1]$, $|\hat{a}_i|$ and $|\hat{a}|$ indicate the total number of encoded patterns from tool response and preliminary answer, we always have $P(\hat{a},\hat{a}_i) \geq 0$. In this proof, for better understanding, we assume a most common case that each word token is encoded to only one pattern, namely no word token exhibits multiple patterns. Thus, $|\hat{a}_i|$ and $|\hat{a}|$ are equal to the length of unencoded sequences of tool response and preliminary answer. $p_{\hat{a}_i}(\cdot) = \frac{C_j^{\hat{a}_i}}{|\hat{a}_i|}$ and $p_{\hat{a}}(\cdot) = \frac{C_j^{\hat{a}}}{|\hat{a}|}$ represent the probability of $j^{th}$ pattern in raw $\hat{a}_i$ and $\hat{a}$.
\begin{align*}
P(\hat{a}, \hat{a}_i) &= \sum_{j\in \{1,\cdots,|\mathcal{S}|\}} \frac{(C_j^{\hat{a}} + \lambda)C_j^{\hat{a}_i}}{(|\hat{a}| + \lambda|\mathcal{S}|)|\hat{a}_i|}\log \frac{1}{\mathcal{P}_j} \\
&= \sum_{j\in \{1,\cdots,|\mathcal{S}|\}} \frac{C_j^{\hat{a}}C_j^{\hat{a}_i} + \lambda C_j^{\hat{a}_i}}{|\hat{a}| |\hat{a}_i| + \lambda|\mathcal{S}||\hat{a}_i|}\log \frac{1}{\mathcal{P}_j}
\end{align*}
If $\lambda = 0$:
\begin{align*}
&P(\hat{a},\hat{a}_i) \\
&= \sum_{j\in \{1,\cdots,|\mathcal{S}|\}} \frac{C_j^{\hat{a}}C_j^{\hat{a}_i}}{|\hat{a}||\hat{a}_i|}\log \frac{1}{\mathcal{P}_j} \\ 
&= \sum_{x\in \{E(\hat{a}_i)\}}\sum_{y\in \{E(\hat{a})\}} p_{\hat{a}_i,\hat{a}}(x,y) \mathbb{I}_{x=y}\log \frac{1}{\mathcal{P}(x)} \\
&= \sum_{x\in \{E(\hat{a}_i)\}}\sum_{y\in \{E(\hat{a})\}} p_{\hat{a}_i}(x)p_{\hat{a}}(y)\mathbb{I}_{x=y}\log \frac{1}{\mathcal{P}(x)} \\
&= \sum_{x\in \{E(\hat{a}_i)\}}p_{\hat{a}_i}(x)\log \frac{1}{\mathcal{P}(x)}\sum_{y\in \{E(\hat{a})\}}p_{\hat{a}}(y)\mathbb{I}_{x=y} \\
&\leq \sum_{x\in \{E(\hat{a}_i)\}}p_{\hat{a}_i}(x)\log \frac{1}{\mathcal{P}(x)} \cdot p_{\hat{a}}(x) \\ &\quad \; (\text{Because it is possible } \{E(\hat{a}_i)\}\cap \{E(\hat{a})\} = \phi)\\
&= \sum_{x\in \{E(\hat{a}_i)\}}p_{\hat{a}_i}(x)p_{\hat{a}}(x)\log \frac{1}{\mathcal{P}(x)} \\
&\leq \sum_{x\in \{E(\hat{a}_i)\}}p_{\hat{a}_i}(x)\log \frac{1}{\mathcal{P}(x)}\\
&= \text{CE}(p_{\hat{a}_i}, \mathcal{P})
\end{align*}
$\text{CE}(p_{\hat{a}_i}, \mathcal{P})$ is the cross-entropy between the pattern distribution of raw tool response and the prior pattern distribution. $\{E(\hat{a}_i)\}$ and $\{E(\hat{a})\}$ are two sets of patterns derived from encoding tool response $\hat{a}_i$ and preliminary answer $\hat{a}$, respectively. $p_{\hat{a}_i,\hat{a}}(x,y)$ is the joint probability of pattern $x$ and $y$ in $\hat{a}_i$ and $\hat{a}$. Because $\hat{a}_i$ and $\hat{a}$ are obtained independently, we can simply write the joint probability as the product of $p_{\hat{a}_i}(x)$ and $p_{\hat{a}}(y)$. $\mathbb{I}_{x=y}$ is the indicator function. $\mathcal{P}(x)$ is the prior probaility of the pattern $x$. Note that unlike $j$ which is an index variable, $x$ and $y$ here are real pattern variables.\\
\\
If $\lambda > 0$: let $\delta \subseteq \{1,2,\cdots,|\mathcal{S}|\}$ such that $C^{\hat{a}}_\alpha > 0$ for $ \alpha \in \delta$ and $C^{\hat{a}}_\beta = 0$ for $\beta\in \{1,2,\cdots,|\mathcal{S}|\} \setminus \delta$.
\begin{align*}
& P(\hat{a},\hat{a}_i) \\
&= \sum_{\alpha\in \delta} \frac{C^{\hat{a}}_\alpha C^{\hat{a}_i}_\alpha + \lambda C^{\hat{a}_i}_\alpha}{|\hat{a}| |\hat{a}_i| + \lambda|\mathcal{S}||\hat{a}_i|}\log \frac{1}{\mathcal{P}_\alpha} \\ &\quad \quad\quad + \sum_{\beta\in \{1,2,\cdots,|\mathcal{S}|\} \setminus \delta}\frac{\lambda C^{\hat{a}_i}_\beta}{|\hat{a}| |\hat{a}_i| + \lambda|\mathcal{S}||\hat{a}_i|}\log \frac{1}{\mathcal{P}_\beta} \\
&= \sum_{\alpha\in \delta} \frac{C^{\hat{a}_i}_\alpha}{|\hat{a}_i|}\cdot \frac{C^{\hat{a}}_\alpha + \lambda }{|\hat{a}| + \lambda|\mathcal{S}|}\log \frac{1}{\mathcal{P}_\alpha} \\ &\quad \quad \quad + \sum_{\beta\in \{1,2,\cdots,|\mathcal{S}|\} \setminus \delta} \frac{C^{\hat{a}_i}_\beta}{|\hat{a}_i|} \cdot \frac{\lambda}{|\hat{a}| + \lambda|\mathcal{S}|}\log \frac{1}{\mathcal{P}_\beta} \\
&\leq \sum_{\alpha\in \delta} \frac{C^{\hat{a}}_\alpha + \lambda}{|\hat{a}| + \lambda|\mathcal{S}|}\log \frac{1}{\mathcal{P}_\alpha}  \\ &\quad \quad \quad + \sum_{\beta\in \{1,2,\cdots,|\mathcal{S}|\} \setminus \delta}\frac{\lambda}{|\hat{a}| + \lambda|\mathcal{S}|}\log \frac{1}{\mathcal{P}_\beta} \\
&= \text{CE}_{\alpha}(\tilde{p}_{\hat{a}}, \mathcal{P}) + \lambda \text{CE}_{\beta}(U(0, |\hat{a}| + \lambda |\mathcal{S}|), \mathcal{P})
\end{align*}
where $U$ is the uniform distribution and $\tilde{p}_{\hat{a}}$ is the smoothed pattern distribution of $\hat{a}$.

\subsection{Pattern Similarity Score Properties}
\label{Appendix: Pattern Similarity Score Properties Illustration}
\begin{itemize}[noitemsep, leftmargin=*]
    \item \emph{Order Insensitive:}  The position of a pattern should not influence the score, as the preliminary answer generated by the \slm{} tends to be disorganized. 
    \item \emph{Length Insensitive:}  The score should not be biased toward the length of tools' responses, as certain tools are inclined to generate longer responses. 
    \item \emph{Pattern Sensitive:} Given the prior distribution $\mathcal{P}$, tools that exhibit rare patterns are more likely to be chosen when the preliminary answer $\hat{a}$ also exhibits those patterns. 
    \item \emph{Pattern Set Size Insensitive:}  The average pattern similarity score should remain consistent for various tool library and pattern set sizes. This property ensures a consistent hyperparameter $\gamma$ (the weight for semantic and pattern scores).
    \item \emph{Commutative:}  $P(\hat{a},\hat{a}_i) = P(\hat{a}_i,\hat{a})$ should be hold for any preliminary answer $\hat{a}$ and tool responses $\hat{a}_i$.
\end{itemize}
\begin{table}[t]
\centering
\begin{tabular}{lrrrr}
\toprule
\textbf{Properties}                   & $\hat{a}$ (Encoded)                                      & $\hat{a}_1$ (Encoded)                                          & $\hat{a}_2$ (Encoded)                                          & \textbf{Result}                              \\ \midrule
Order Insensitive            & \texttt{ene}                  & \texttt{ene}                  & \texttt{een}                  & $P(\hat{a}, \hat{a}_1) = P(\hat{a},\hat{a}_2)$  \\ 
Length Insensitive           & \texttt{eee}                  & \texttt{en}                    & \texttt{enenen}             & $P(\hat{a}, \hat{a}_1) = P(\hat{a},\hat{a}_2)$  \\ 
Pattern Sensitive            & \texttt{ene}                  & \texttt{ene}                  & \texttt{enn}                  & $P(\hat{a}, \hat{a}_1) < P(\hat{a},\hat{a}_2)$  \\ 
Commutative &\texttt{ene} & \texttt{eee}& \texttt{nnn} & $P(\hat{a}, \hat{a}_1) = P(\hat{a}_1, \hat{a})$ \\ \bottomrule
\end{tabular}
\caption{Examples illustrating the four essential properties of pattern similarity scores}
\label{table:pattern score properties}
\end{table}
\autoref{table:pattern score properties} gives illustrative examples for the pattern similarity score. \texttt{e} and \texttt{n} denote English token pattern and number pattern. The less frequency of numbers ``\texttt{n}'' in real corpus compared to English tokens ``\texttt{e}'' results in a smaller prior probability $\mathcal{P}(n) < \mathcal{P}(e)$, leading to the result in the Pattern Sensitive row. In other words, with the same length, tool response $\hat{a}_2$ containing more rare patterns which also exhibit in the preliminary answer $\hat{a}$ would have higher pattern similarity score. 

The Pattern Set Size Insensitive property also holds because the denominator $(|\hat{a}| + \lambda|\mathcal{S}|)|\hat{a}_i|$ is insensitive to the $\Delta |\mathcal{S}|$, given that $|\hat{a}| \gg |\mathcal{S}|$ and $\lambda$ is typically small. Therefore, as long as the tool response or preliminary answer does not exhibit the given patterns, namely $C^{\hat{a}}_j = 0$ or $C^{\hat{a}_i}_j = 0$, $P(\hat{a}, \hat{a}_i)$ would not significantly change regardless of the size of $|\mathcal{S}|$.

To prove the length-insensitive property, we have to first assume tool responses $\hat{a}_1$ and $\hat{a}_2$ share the same pattern probability distribution. Namely, we have
$$
\frac{C^{\hat{a}_1}_j}{|\hat{a}_1|} = \frac{C^{\hat{a}_2}_j}{|\hat{a}_2|},\; \forall j \in \{1,2,\cdots,|\mathcal{S}|\}
$$
Then the comparison of pattern similarity scores for these two tools is only determined by the preliminary answer $\hat{a}$, pattern set size $|\mathcal{S}|$ and $\lambda$, with no sensitivity to the length of tool responses.

\subsection{Mock Pattern}
\label{Appendix: Mock pattern}
When dealing with a large tool library, iterating through all tools for true responses is inefficient and some tools may not have textual responses to encode. Conversely, through the utilization of pattern scores, we can set certain tools to generate mock responses with corresponding mock patterns during the tool grounding process, eliminating the requirement for actual execution, thereby reducing the \name's time complexity and generalizing it to various types of tools. In the experiment section \S\ref{section:experiment}, we test the efficiency and generalizability of mock patterns for tool grounding by adding \sleep{} and \map{} to the tool library.

\clearpage
\section{Implementation Details}
\label{Appendix:implementation details}
\subsection{Hyperparameters}
\label{Appendix:hyperparameters}
To avoid bias toward to pattern similarity score, we use add-one smoothing and set $\lambda = 1$. Additionally, based on our experiment, we observed that the mean of pattern similarity score is consistently three times greater than the mean of the semantic score. In order to achieve a proper balance between these two scores, we set $\gamma=0.75$ throughout the entire experiment. 

\subsection{Patterns}
\label{Appendix: patterns setup}
For 4 tools experiments, we use the following four patterns: $\mathcal{S}$ = \{English token pattern: \texttt{e}, non-ASCII token pattern: \texttt{f}, number pattern: \texttt{n}, symbol pattern: \texttt{s}\}. Because we believe these four basic patterns could cover a lot of language tools. Based on their frequency in the real corpus, we set their prior probabilities as follows: $\mathcal{P}$ = \{\texttt{e}: 0.78, \texttt{f}: 0.18, \texttt{n}: 0.05, \texttt{s}: 0.02\}.

For generalization experiments where the tool library size varies between 4 to 10, we consistently use the following prior pattern distribution: $\mathcal{P}$ = \{\texttt{e}: 0.75, \texttt{f}: 0.15, \texttt{n}: 0.02, \texttt{s}: 0.02, Sleep Pattern: 0.02, Move pattern: 0.02, Time pattern: 0.02\}. 
\subsection{Models}
\label{Appendix: Model Preparation}
\begin{itemize}[leftmargin=0.1in]
    \item \gptj{} is from \url{https://huggingface.co/EleutherAI/gpt-j-6b}
    \item \gptNeo{} is from \url{https://huggingface.co/EleutherAI/gpt-neo-1.3B}
    \item \miniLM{} is from \url{https://huggingface.co/sentence-transformers/all-MiniLM-L6-v2}
    \item \mpnet{} is from \url{https://huggingface.co/sentence-transformers/all-mpnet-base-v2}
\end{itemize}

\subsection{Tasks and Datasets} 
\label{Appendix:datasets}
We use the following 12 datasets from 6 downstream tasks in our main experiment, plus 2 extra datasets (SQuAD~\citep{rajpurkar2016squad} and Trex~\citep{elsahar-etal-2018-rex}) in~\autoref{table:Toolformer comparison}. To keep the evaluation costs manageable, we use 1K instances per dataset.
\begin{itemize}[leftmargin=0.1in]
    \item \textbf{Arithmetic (Arithm)}: 
    We evaluate on ASDiv~\citep{miao-etal-2020-diverse}, GSM8K~\citep{cobbe2021gsm8k} and SVAMP~\citep{patel-etal-2021-nlp} datastes. Given the arithmetic nature of these datasets, we expect successful grounding in \calc{} tool should improve their performance.  
    \item \textbf{Machine Translation (MT)}: We use IWSLT-2017~\citep{cettolo-etal-2017-overview} dataset to evaluate the utility of successful grounding to the \mt{} tool. The input data consists of an English prompt and a non-English context in Simplified Chinese, Arabic, German, French, Japanese, or Korean. We utilize diverse English prompts for English translation requests (e.g., ``How do you say ... in English'', ``Speak ... to English'', etc.). We sample 1K instances for each source language.
    \item \textbf{Open-domain QA (ODQA)}: We experiment with NQ-Open~\citep{lee2019latent}, WebQS~\citep{berant2013semantic}, and TriviaQA~\citep{joshi-etal-2017-triviaqa}, since open-domain questions require external knowledge, successful grounding of these tasks to \wiki{} tool improve their performance.
    \item \textbf{Commensense QA (CSQA)}: To investigate the benefit of utilizing the \qa{} tool, we evaluate all baselines on CommonsenseQA~\citep{talmor-etal-2019-commonsenseqa}, COPA~\citep{roemmele2011choice}, and SocialIQA~\citep{sap-etal-2019-social}. Those datasets require the model to perform commonsense reasoning for a given context and select the answer from a variety of choices.
    \item \textbf{Multilingual QA (MLQA)}: MLQA~\citep{lewis2020mlqa} is a hard multilingual question-answering benchmark, expecting \mlqa{} to tackle such problem. Each instance includes an English context and a query presented in Arabic, German, Spanish, Hindi, Vietnamese, or Chinese. We randomly sample 1K instances for each language.
    \item \textbf{Timezone Conversion}: we create this dataset programmatically by iterating over all combinations of time zones, randomly-generated numbers which are verbalized into the natural language via real querying scenarios. Specifically, we set 5 querying templates and 3 time formats, combining them with randomly selected timezones to construct the dataset.
    Here are two examples:
    \begin{mdframed}
    \footnotesize
    \textit{My friend is in Cordoba, and I am in Madeira. If it is 2016-07-14 08:24:07 here, what time is it there?} \\
    \\
    \textit{I want to make a call to someone. He is in Johannesburg, and I am in Pitcairn. If it is May 16 2023 10:31:14AM here, what time is it there?}
    \end{mdframed}

    \normalsize
    Successfully grounding to the \tz{} should improve the performance of this task.
\end{itemize}

\paragraph{Evaluation Metrics.} 
For the Arithmetic task, we convert all English numerals to their numerical equivalents and then pick the last number as the answer.\footnote{
For zero-shot or few-shot baseline the overall answer typically appears after the rationales. 
} These are not needed when using \calc{} tool, as it always outputs a single number. Ultimately, we compute an exact match accuracy between the resulting numbers and gold answers.
For ODQA and MLQA tasks, following ~\citep{schick2023toolformer},  we verify if the generated output contains the gold answer. 
For the CSQA task, we compute the accuracy as the ratio of accurately selected outputs.
For the MT task, the translation quality is evaluated using a BLEU (as percentage). 

\clearpage
\section{Tools}
\label{Appendix:tools}
We prioritize two factors for choosing tools: 1) whether they could compete with others 2) whether their function is naturally beyond the capability of any \llm{}. 
\subsection{Basic Tools}
Description and usage prompts for each basic tool are provided in \autoref{table:Prompt}

\begin{itemize}[leftmargin=0.1in]
    \item \qa{}: Our question-answering system is based on an external language model specially trained for answering questions. We utilize ChatGPT in our experiment, renowned for its performance in comprehending and reasoning with human language.
    \item \calc{}: The \calc{} is built from the Python built-in function \textit{eval}, which supports four fundamental arithmetic operations with priorities that can be specified using brackets. The output is rounded to three decimal places.
    \item \mt{}: The core of our machine translation tool is the Google Translate API. It accepts two input arguments: the text to be translated and the target language.
    \item \wiki{}: The last basic tool employed in our experiment is the Wikipedia Search (Wikisearch) engine. It returns wiki paragraphs in response to queries. This tool advances models by supplying external factual knowledge and its returned output is more formal and informative than that of \qa{}. In our experiment, we use ColBERTv2~\citep{santhanam2022colbertv2} as the search retriever to index relevant information.
\end{itemize}
\label{Appendix:basic tools}

\subsection{Novel Tools}
For the selection of novel tools, we follow these two factors: whether they could compete with existing tools or whether their function is naturally beyond the capability of any \llm{}. Consequently, we add the following six tools:

\begin{itemize}[leftmargin=0.1in]
    \item \logcalc{} and \expcalc{}: These two tools aim to solve logarithm and exponential problems and serve as competitors to the \calc{} tool.
    \item \mlqa{}: We compose \mt{} and \qa{} tools to form the \mlqa{} pipeline. It involves two steps: translating the query to the target language using \mt{}, and passing the context and translated query to the \qa{} to find the final answer.
    \item \tz{}: This tool is implemented by the Python \textit{pytz} library. It converts a time from one time zone to another. Such a task is also solvable by the QA tool but not accurately. Therefore, we want to assess the success rate of grounding the most appropriate tools for such endeavors.
    \item \sleep{}: This tool suspends the entire program for a specified duration. This tool is intended to test the mock response functionality for our system. We do not expect the program to sleep during the tool grounding procedure; a mocked response is sufficient. However, once selected, this tool should perform its intended function.
    \item \map{}: This tool instructs a robot to move a specified distance in a chosen direction. Similarly to \sleep{}, this tool is used for testing the mock response for grounding tools with non-textual outputs. During the grounding process, its returned response is a mock text: ``\texttt{Robot is moving forward for \{\} meters}''. 
\end{itemize}
\label{Appendix: novel tools}

\clearpage
\section{Downstream Performance}
\label{Appendix: Performance Experiment}
Results for \gptThree{} baselines can be seen in \autoref{table:GPT3 acc}. For MT and Commensense QA tasks, even the few-shot performance is lower than zero-shot, we hypothesize that this is because the \gptThree{} model has seen those datasets during the pretraining and memorized them.

A comparison of \name{} with ART~\citep{paranjape2023art} on Arithmetic and Open-domain QA tasks is provided in \autoref{table:ART comparison}. The downstream accuracy of \name{}-augmented \gptThree{} is only slightly higher than those of ART-augmented \gptThree{}, because according to \autoref{table:generalization test}, ART achieves at least $90\%$ grounding accuracy on most Arithmetic and Open-domain QA datasets. However, it is worth noting that ART requires in-domain demonstration for each task/dataset while \name{} does not.
\begin{table*}[t]
\small
\centering
\begin{tabular}{lcccccc}
\toprule
        \textbf{Models}              & \textbf{ASDiv} & \textbf{GSM8K} & \textbf{SVAMP} & \textbf{NQ-Open} & \textbf{WebQA} & \textbf{TriviaQA} \\ \midrule   
\textbf{ART$_{\text{cs}}$ (\miniLM/\gptDavinciThree)}      & \textbf{86.7}           & 69.7           & 77.3           & 56.7           & 17.7              & 61.0            \\
\textbf{\name{} (\gptNeo/\gptDavinciThree)}     & 74.9 (-11.8)           & \textbf{71.1 (+1.4)}           & \textbf{79.9 (+2.6)}           & 53.8 (-2.9)           & \textbf{23.6 (+5.9)}              & \textbf{62.5 (+1.5)}              \\ \bottomrule
\end{tabular}
\caption{Comparing \name{} with ART~\citep{paranjape2023art} on Arithmetic and Open-domain QA tasks}
\label{table:ART comparison}
\end{table*}

\begin{table*}[t]
\small
\centering
\begin{tabular}{@{}l*4c@{}}
\toprule
\multirow{4}{*}{\begin{tabular}[c]{@{}l@{}}Algorithm $\rightarrow$ \\ Grounding Model $\rightarrow$ \\Execution Model $\rightarrow$\\Datasets $\downarrow$\end{tabular}} & \multirow{3}{*}{\begin{tabular}[c]{@{}c@{}}Zero-shot \\  $\rule[0.5ex]{0.5em}{0.4pt}$ \\ \gptThree \end{tabular}} & \multirow{3}{*}{\begin{tabular}[c]{@{}c@{}} Few-shot \\  $\rule[0.5ex]{0.5em}{0.4pt}$  \\ \gptThree\end{tabular}} & \multirow{3}{*}{\begin{tabular}[c]{@{}c@{}}ART$^\ast_{\text{llm}}$ \\ \gptNeo \\ \gptThree \end{tabular}}& \multirow{3}{*}{\begin{tabular}[c]{@{}c@{}}\name\\ \gptNeo \\ \gptThree \end{tabular}}  \\
                          &&&&\\
                          &&&&\\
                          &&&&\\
                         \midrule
ASDiv              & 78.7
      & 75.3             & 37.0        & 74.9          \\
GSM8K        & 62.4
             & 69.9             & 14.7        & 71.1            \\
SVAMP        & 75.4
           & 73.7             & 21.3        & 79.9          \\ \hdashline\noalign{\vskip 0.5ex}
\arrowdownright \footnotesize \textbf{Average (Arithm)}                  & 72.2 & 73.0  & 24.3             & \textbf{75.3}                \\ \midrule
IWSLT(cn)           & 43.1
    & 30.1             & 19.2       & 39.2              \\
IWSLT(ar)                 & 47.2 & 41.1             & 16.1       & 41.8            \\
IWSLT(de)               & 51.6  & 40.8             & 25.0        & 51.0               \\
IWSLT(fr)        & 55.8  & 42.7             & 25.9        & 55.0               \\
IWSLT(ja)       & 31.4
        & 28.6             & 13.2        & 28.8                \\
IWSLT(ko)     & 37.9           & 31.3             & 16.5       & 36.5               \\ \hdashline\noalign{\vskip 0.5ex}
\arrowdownright \footnotesize \textbf{Average (MT)}         & \textbf{44.5}        & 35.8             & 19.3        & 42.0        \\ \midrule
NQ-Open      & 58.0
     & 66.1             & 24.0        & 53.8      \\
WebQS        & 24.9
     & 28.1             & 11.2        & 23.6         \\
TriviaQA       & 54.9
         & 70.4             & 24.3        & 62.5            \\\hdashline\noalign{\vskip 0.5ex}
\arrowdownright \footnotesize \textbf{Average (ODQA)}     & 45.9 & \textbf{54.9}      & 19.8             & 46.6               \\ \midrule
CSQA        & 74.7            & 75.6             & 5.0         & 70.1               \\
COPA            & 45.5
      & 33.7             & 0.3       & 36.7            \\
SocialIQA        & 56.8
    & 64.8             & 1.2            & 59.5            \\ \hdashline\noalign{\vskip 0.5ex}
\arrowdownright \footnotesize \textbf{Average (CSQA)}      & \textbf{59.0}  & 58.0             & 2.2        & 55.4        \\ \bottomrule
\end{tabular}
\caption{Downstream task performance result. Evidently, \textbf{\name{}-augmented \gptThree{} achieves competitive results with \gptThree{} few-shot and ART}, both of which provided with task-specific demonstrations for solutions.}
\label{table:GPT3 acc}
\end{table*}
\clearpage
\section{Grounding Performance}
\label{Appendix:grounding experiment}
\begin{figure*}[t]
    \centering
    \includegraphics[width=0.49\textwidth]{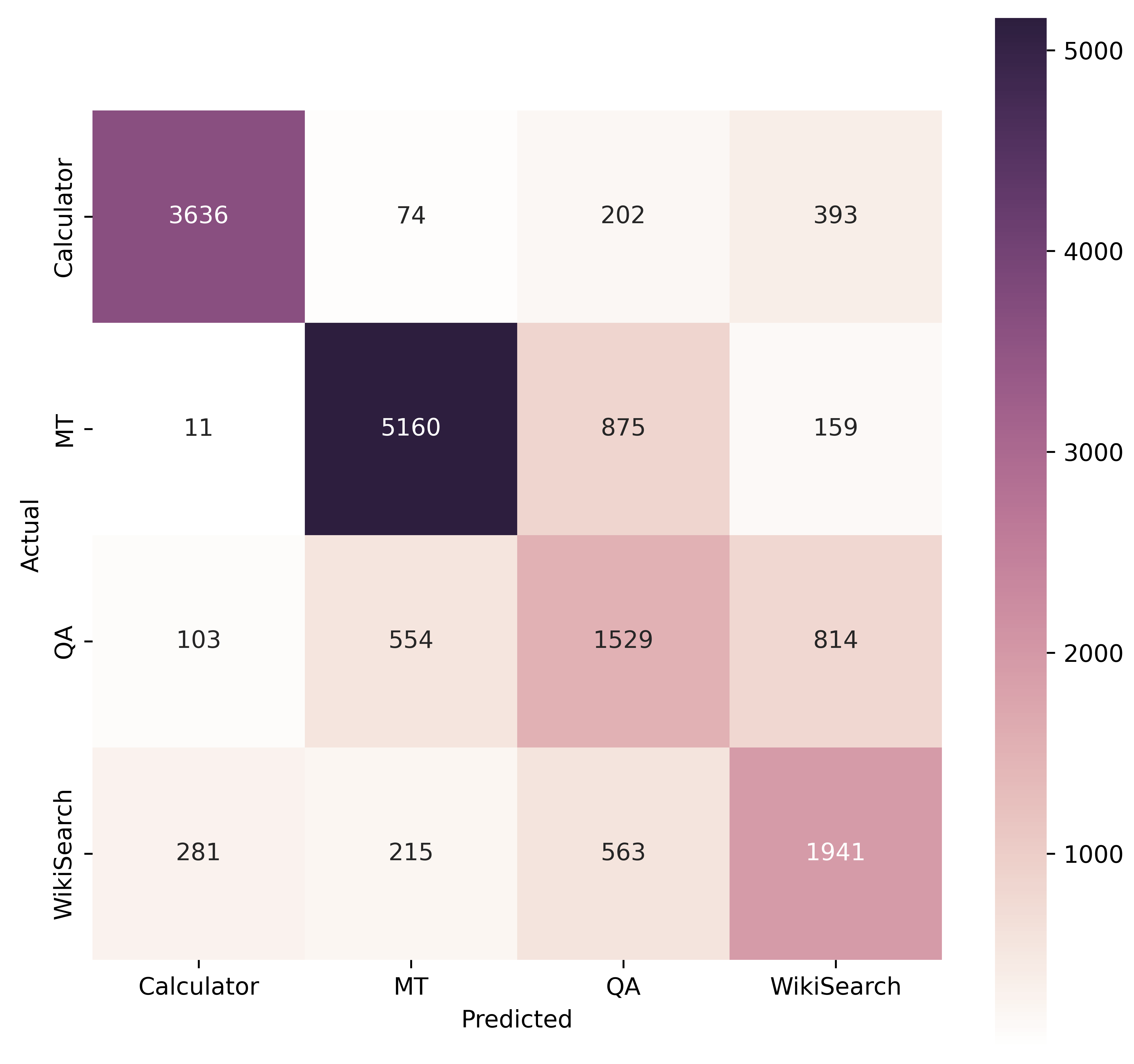}
    \caption{Confusion matrix of grounding results of four basic tools. Each number represents the number of examples being grounded to the tool.}
    \label{fig:confusion matrix}
\end{figure*}


\begin{table}[t]
\centering
\begin{tabular}{llcc}
\toprule
\textbf{Dataset (w/10 Tools)}    & \textbf{Target Tool} & \name   & ART$^\ast_{\text{cs}}$    \\ \midrule
Average (Arithm) & \tt{Cal}   & 74.0 & 97.2 \\
Average (MT)   & \mt{}    & \textbf{80.5} & 78.5 \\
Average (ODQA)  & \wiki{}   & \textbf{40.7} & 21.1 \\
Average (CSQA)  & \qa{}    & \textbf{33.4} & 22.8 \\
Average (MLQA)   & \tt{MLQA}  & \textbf{54.4} & 17.6 \\
Timezone Conversion & \tt{TZ Conveter} & \textbf{96.4} & 95.0 \\ \bottomrule
\end{tabular}
\caption{Tool grounding accuracy for 6 downstream tasks with a 10-tools library (\S\ref{subsec:grounding result}). \textbf{\name{} using \gptNeo{} outperforms ART$^\ast_{\text{cs}}$ using \mpnet{} with the cosine similarity strategy on 5 out of 6 tasks.}
}
\label{table: 10 tools ratio}
\end{table}

\begin{table}[t]
\small
\centering
\begin{tabular}{@{}l*3c@{}}
\toprule
Algorithm $\rightarrow$ & BM25 & KNN & \name{}(\gptNeo)  \\
Datasets $\downarrow$ &&&\\
                         \midrule
ASDiv              & 0.5 
      & 66.5            & 83.1           \\
GSM8K        & 0.2
             & 58.7             & 83.0            \\
SVAMP        & 1.2
           & 75.1             & 89.0       \\ \hdashline\noalign{\vskip 0.5ex}
\arrowdownright \footnotesize \textbf{Average (Arithm)}                  & 0.6 & 66.8  & \textbf{85.0}                   \\ \midrule
IWSLT(cn)           & $\rule[0.5ex]{0.5em}{0.4pt}$
    & 100             & 84.1                \\
IWSLT(ar)                 & $\rule[0.5ex]{0.5em}{0.4pt}$ & 99.7             &   66.6               \\
IWSLT(de)               & $\rule[0.5ex]{0.5em}{0.4pt}$ & 80.0             &    96.9               \\
IWSLT(fr)        & $\rule[0.5ex]{0.5em}{0.4pt}$  & 84.6             & 96.6                   \\
IWSLT(ja)       & $\rule[0.5ex]{0.5em}{0.4pt}$
        & 100             & 72.4                   \\
IWSLT(ko)     & $\rule[0.5ex]{0.5em}{0.4pt}$          & 100             & 82.2                  \\ \hdashline\noalign{\vskip 0.5ex}
\arrowdownright \footnotesize \textbf{Average (MT)}         & $\rule[0.5ex]{0.5em}{0.4pt}$        & \textbf{94.1}             & 83.1           \\ \midrule
NQ-Open      & 76.2
     & 73.4             & 63.0            \\
WebQS        & 45.4
     & 55.9            & 65.6             \\
TriviaQA       & 62.5
         & 83.3             & 54.3                \\\hdashline\noalign{\vskip 0.5ex}
\arrowdownright \footnotesize \textbf{Average (ODQA)}     & 61.4 & \textbf{70.9}     & 61.0                \\ \midrule
CSQA        & 24.6	           & 75.8	           & 77.1                   \\
COPA            & 0.6	
      &32.9           & 41.3            \\
SocialIQA        & 14.5 
    &	57.2              & 75.7         \\ \hdashline\noalign{\vskip 0.5ex}
\arrowdownright \footnotesize \textbf{Average (CSQA)}      & 13.2  & 55.3             & \textbf{64.7}          \\ \bottomrule
\end{tabular}
\caption{Tool grounding accuracy for 4 downstream tasks with a 4-tools library ("$\rule[0.5ex]{0.5em}{0.4pt}$" denotes $0$). \textbf{\name{} with \gptNeo{} consistently achieves high grounding performance compared to BM25 and KNN.}}
\label{table:retrieval baselines}
\end{table}
According to \autoref{fig:confusion matrix}, it is clear that \calc{} and \mt{} tools have no strong competitors on Arithmetic and MT tasks, while \qa{} and \wiki{} tools are more likely to compete with each other on CommonsenseQA and Open-domain QA tasks. This is due to the functional overlap of these two tools on open-ended NLP tasks.
\paragraph{\name{} is more generalizable than retrieval-based baselines} We compare \name{} with two retrieval-based baselines: Okapi BM25~\citep{robertson1995okapi} and KNN~\citep{fix1989discriminatory} with 50 training examples for each tool under the 4-tools library. Like \name{}, BM25 is a general-purpose approach that does not need supervision. However, from \autoref{table:retrieval baselines}, the grounding accuracy of BM25 is smaller than \name{}'s (\gptNeo{} version) on 13/15 datasets. All MT tasks get a $0\%$ accuracy from BM25 since their inputs contain non-ASCII tokens, which are not accounted for in the description of the MT tool. Although the performance of KNN is generally higher than \name{} on MT and Open-domain tasks, it requires training and is easily overfitting, which hinders its generalizability to low-resource tasks that utilize novel tools without sufficient labeled data.

\paragraph{\name{} is generalizable to novel tasks} We further evaluate \name's generalizability to novel tasks using MLQA~\citep{lewis2020mlqa} and Timezone Conversion datasets. From \autoref{table: 10 tools ratio}, \name{} achieves $54.4\%$ and $96.4\%$ grounding accuracy on these two novel tasks with a 10-tools library. It outperforms ART$^\ast_{\text{cs}}$ on 5 out of 6 tasks, revealing its strong generalizability to both large tool libraries and novel tasks.

\section{Ablation Study}
\label{Appendix: Ablation Study}

\begin{figure*}[t]
    \centering
    \includegraphics[width=0.48\textwidth]{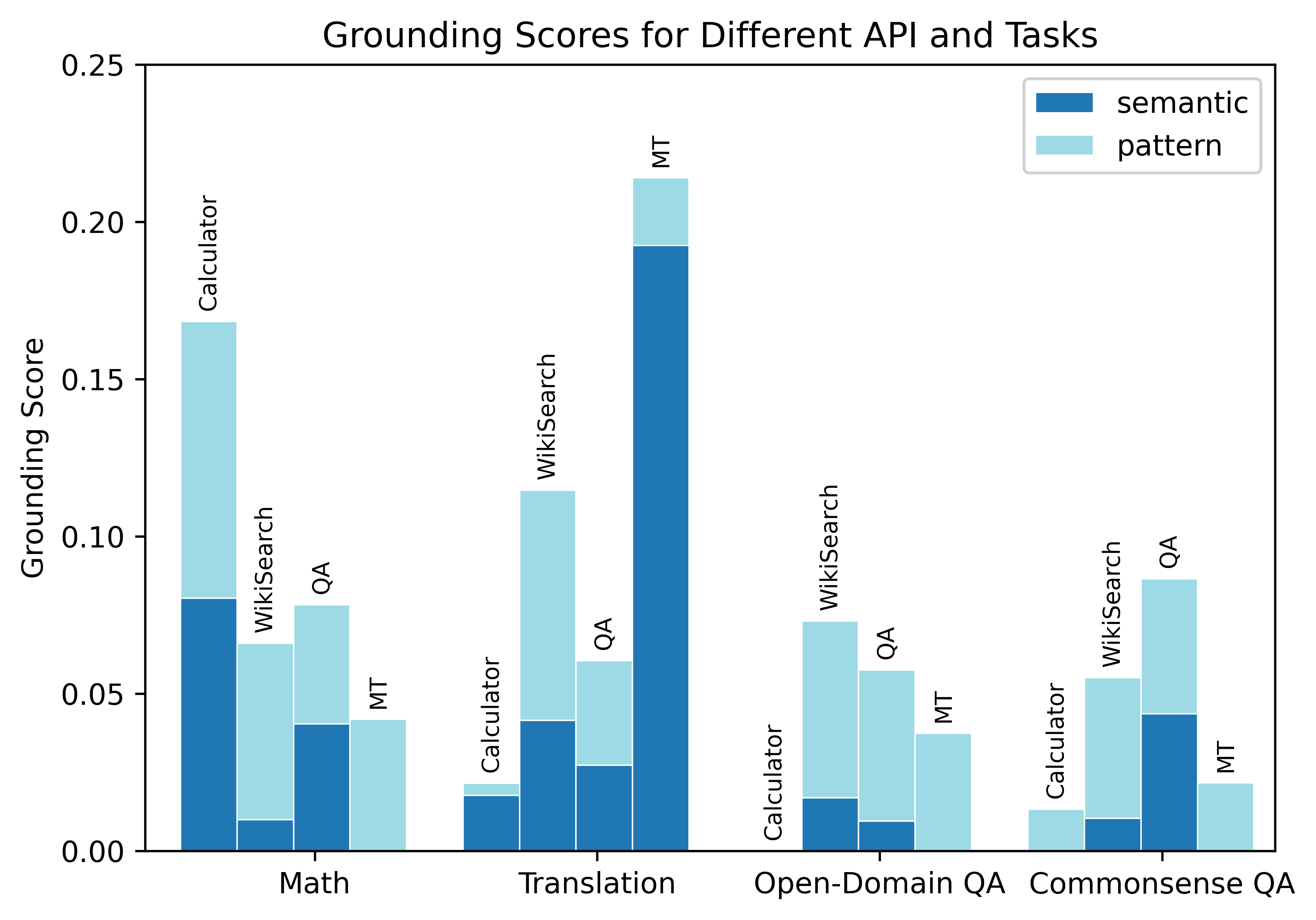}
    \caption{The average similarity scores for different tasks and tools. Clearly, \textbf{the semantic and pattern scores (already weighted by $\gamma$) collaboratively and accurately identify tools for the four basic tasks.}}
    \label{fig:apifilter}
\end{figure*}

For the 4-tools library, we plot the average final grounding score for each task and tool in \autoref{fig:apifilter}. Notably, neither the semantic nor the pattern similarity score dominates the query-tool grounding on most tasks, but they collaborate with each other to correctly identify the tools. 

\clearpage
\section{\name{} Augmented Chatbot}
\label{Appendix:chatbot}
Because \name{} does not require extra task-specific demonstrations, one of its practical applications is that it can be integrated into any chatbot. To validate it, we create a \name{} augmented chatbot using ChatGPT as the execution \llm{} and conduct a survey experiment.

\begin{figure*}[t]
    \centering
    \includegraphics[width=\textwidth]{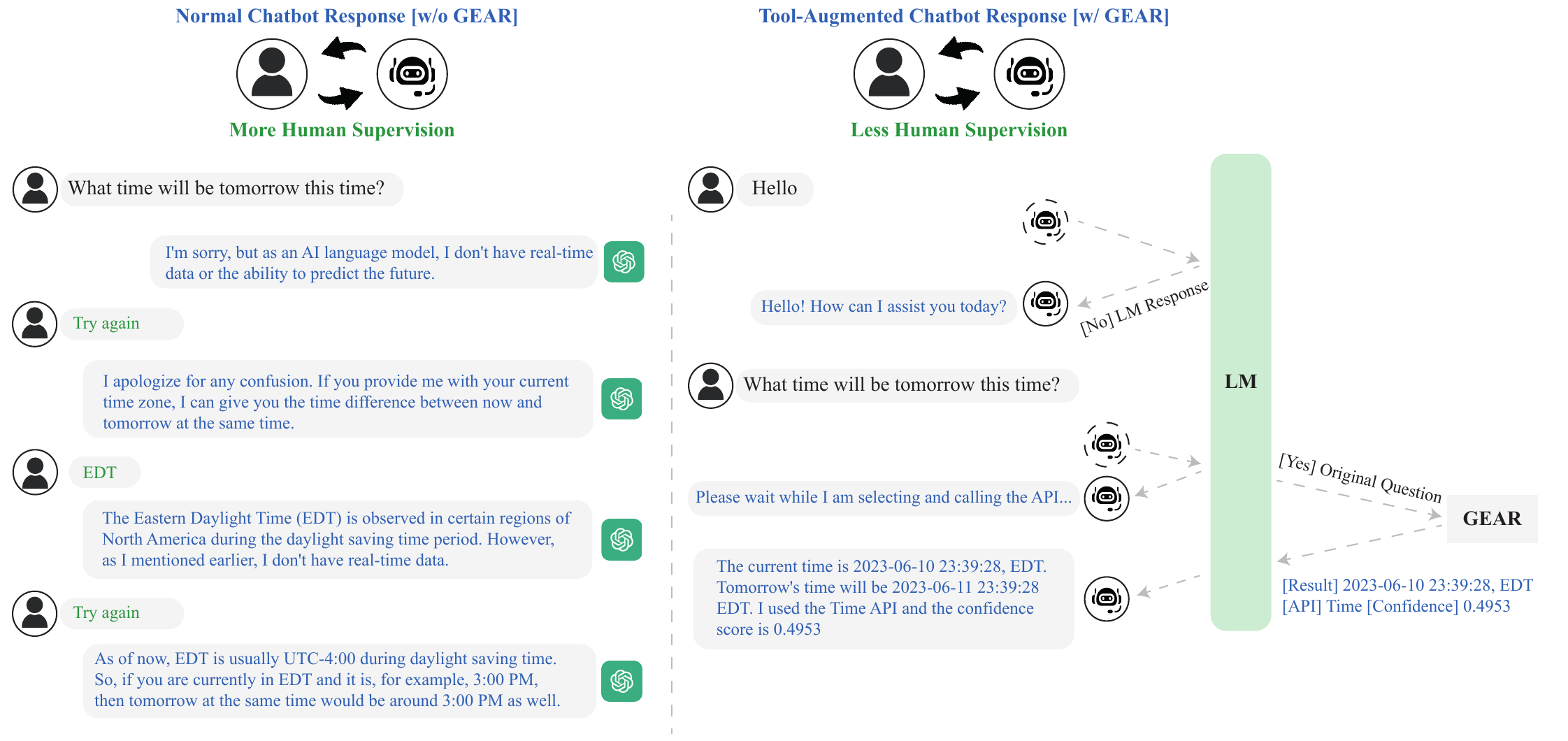}
    \caption{A comparison between the performance of ChatGPT and \name{} augmented chatbot. \name{} requires minimal human supervision, excels in numerous tool-solvable tasks, and offers interpretable confidence scores for users.}
    \label{chatbot}
\end{figure*}

\begin{figure*}[t]
\centering
\begin{subfigure}[b]{\linewidth}
   \includegraphics[width=1\linewidth]{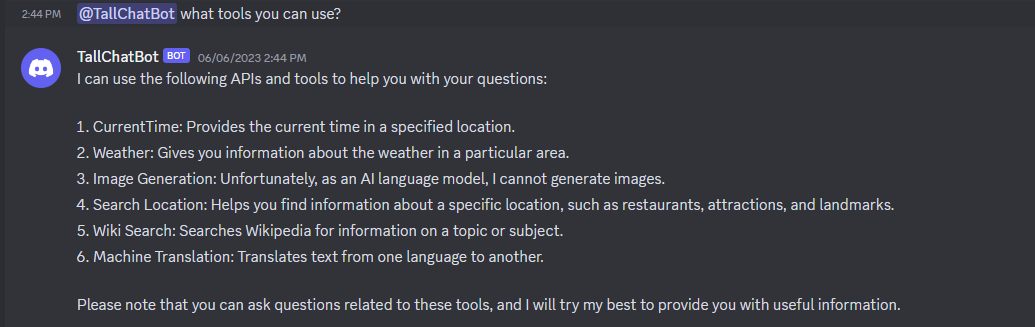}
   \caption{\name{} augmented chatbot screenshot illustrating its tool library.}
\end{subfigure}

\begin{subfigure}[b]{\linewidth}
   \includegraphics[width=1\linewidth]{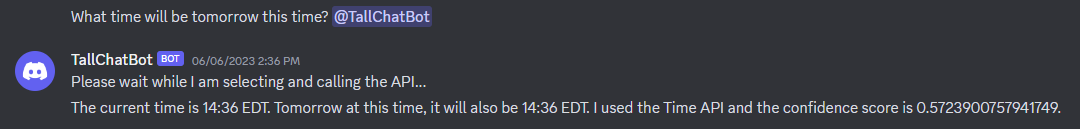}
   \caption{\name{} augmented chatbot screenshot of using the Time tool.}
\end{subfigure}

\begin{subfigure}[b]{\linewidth}
   \includegraphics[width=1\linewidth]{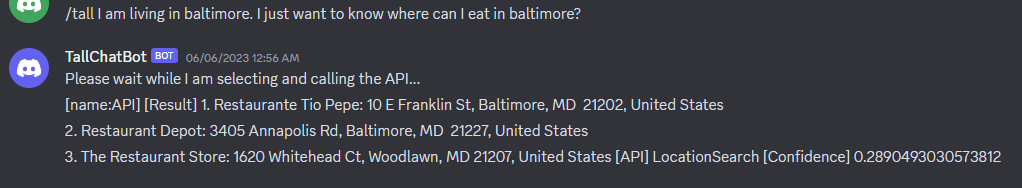}
   \caption{\name{} augmented chatbot screenshot of using the Location Search tool.}
\end{subfigure}
\caption{Screenshots of \name{} augmented chatbot using various tools. Using the command \texttt{/\name} to ask \name{} chatbot to output tool response directly without going through the ChatGPT. While the command \texttt{@TallChatBot} enables a normal conversation where \name{} interacts with ChatGPT to provide more human-readable answers.}
\label{fig:chatbot}
\end{figure*}

\autoref{chatbot} illustrates the differences between \name-augmented chatbot and a normal chatbot and how \name{} interacts with a LM in a dialogue setting. For each user query, we first prompt the LM to determine if a tool usage is necessary. If true, the original query will be sent directly to \name, and \name{} will return the response from the selected tool as well as the tool name and confidence score for selecting that tool. This information is then processed by the LM to generate a more natural, user-friendly response. \autoref{fig:chatbot} provides examples of how our \name{} augmented chatbot works.
We equip it with the following six tools: Weather Search, Location Search, Image Generation, Current Time-Timezone Converter Pipeline, Wikipedia Search and Machine Translation.

We surveyed 50 individuals about the use of our \name{}-augmented chatbot. The evaluators first use ChatGPT-based chatbot for two weeks, then switch to a \name{}-augmented chatbot for the next two weeks. After fully experiencing these two chatbots, they are asked to complete the survey (\autoref{table:survery questions}) which contains four types of questions regarding tool grounding performance and final answer quality. Participants are unpaid and their feedback is unmodified.

The survey reveals that $76\%$ of users agree that integrating tool usages makes the chatbot more useful and fascinating, and more than $90\%$ of queries grounded correct tools. Image generation and weather search tools are the most popular tools among the 6 tools, with more than $50\%$ of users employing them to solve problems. Regarding response quality, our survey indicates that an average of $78.4\%$ of questions are answered to the user's satisfaction, a $16.9\%$ increase in satisfaction compared to the previous chatbot that lacks the tool utilization functionality. The Current Time-Time Zone Convertor Pipeline has the highest accuracy, at $100\%$, while the Machine Translation tool has the lowest quality, with a satisfaction ratio of only $50.5\%$. We infer that the performance of the Google Translate API may not be adequate to satisfy the needs of our evaluators, given that most of them are translating extremely complex sentences between English, Japanese, and Chinese. 

In summary, \name{} substantially improves users' experience on ChatGPT, and it also has excellent generalization capabilities to novel tools. Note that these novel tools lack training data, but with \name{} and just a few words of tool description and usage examples, they can be easily integrated into a chatbot to provide precise and reliable answers to users.

\begin{table}[t]
\small
\centering
\begin{tabular}{p{0.7\textwidth}p{0.15\textwidth}p{0.15\textwidth}}
\toprule
Survery Question                                                                     & Question Type & Answer \\ \midrule
How would you rate your overall experience with \name{}-augmented chatbot?   & rating scale  & 0-10   \\&&\\
Do you think \name{}-augmented chatbot has become smarter compared to the previous version? &  rating scale             &  0-10      \\&&\\
Do you think \name{}-augmented chatbot has become more helpful than the previous one?               &   rating scale            &  0-10      \\&&\\
How accurate do you think the answers of the older bot are?                                                                           &       rating scale            &  0-10      \\&&\\
How accurate do you think the answers of the new version bot are?                                                                            &          rating scale            &  0-10   \\&&\\
Have you noticed that the chatbot is using external tools to help you?    & Likert scales             & yes or no       \\&&\\
How would you rate the chatbot's accuracy in choosing the right tool to answer your query?
                                                                            &  rating scale             &  0-10      \\&&\\
Can you recall a situation where the chatbot chose the wrong tool for your query? If so, please describe it briefly.
                                                                            &   open-ended            &   open-ended    \\ &&\\
Have you ever instructed the chatbot to use a different tool for your query, or did the chatbot automatically choose a different tool because you weren't satisfied with the results? &  Likert scales & yes or no \\&&\\
Will the chatbot be able to switch to the right tool based on your instructions? & Likert scales & yes or no \\&&\\
When a chatbot uses an external tool, how would you rate its response accuracy? & rating scale & 0-10 \\&&\\
Can you recall any instances where the chatbot used external tools to produce output errors or didn't meet your expectations? If so, please describe it briefly & open-ended & open-ended \\&&\\
What tools of chatbots have you used? & multiple-choice & multiple-choice \\&&\\
How would you rate the accuracy of the output generated by the chatbot using the Time tool? & rating scale             &  0-10 \\&&\\
How would you rate the accuracy of the output generated by the chatbot using the Wikisearch tool? & rating scale             &  0-10 \\&&\\
How would you rate the accuracy of the output generated by the chatbot using the Weather Lookup tool? & rating scale             &  0-10 \\&&\\
How would you rate the accuracy of the output generated by the chatbot using the Location Search tool? & rating scale             &  0-10 \\&&\\
How would you rate the accuracy of the output generated by the chatbot using the Image Generation tool? & rating scale             &  0-10 \\&&\\
How would you rate the accuracy of the output generated by the chatbot using the Machine Translation tool? & rating scale             &  0-10 \\&&\\                                                                            
Please provide any additional feedback or suggestions you have for improving \name{}-augmented chatbot performance. & open-ended & open-ended \\&&\\
Overall Score you want give to the \name{}-augmented chatbot & rating scale & 0-100 \\  \bottomrule
\end{tabular}
\caption{Survey Questions}
\label{table:survery questions}
\end{table}

\clearpage
\section{Prompts}
\label{Appendix:prompts}
\newcommand{\style}{\tt\footnotesize}

\autoref{table:apicall} provides examples of API calls and outputs for each tool

\noindent \autoref{table:few-shot demonstrations} shows task-specific demonstrations used for the few-shot baseline in the experiment

\noindent \autoref{table:Prompt} presents the description and usage example of each basic tool.

\begin{CJK*}{UTF8}{gbsn}
\begin{table*}[h]
\begin{tabular}{p{0.15\textwidth}p{0.55\textwidth}p{0.3\textwidth}}
\toprule
\textbf{Tool} & \textbf{Example API Call} & \textbf{Example Output} \\ \midrule
Question Answering  & QA("What century did the Normans first gain their separate identity?") & The Normans first gained their separate identity in the 11th century. \\&&\\
Calculator & Calculator(2 + 4) & 6 \\&&\\
Machine Translation & MT("太多东西要在这18分钟内讲述了。", "en") & There are too many things to be described in this 18 minutes. \\&&\\
Wikipedia Search    & WikiSearch("Lord Of The Flies")  & Lord of the Flies (song) "Lord of the Flies" is an Iron Maiden single and second track on their 1995 album "The X Factor". \\&&\\
Multilingual QA & MultilingualQA("question: 《街机游戏街头霸王II》的游戏机上有多少用于控制角色的圆形物体？ context: For example, the six button layout of the arcade games Street Fighter II or Mortal Kombat cannot be comfortably emulated on a console joypad, so licensed home arcade sticks for these games have been manufactured for home consoles and PCs.") & Six \\&&\\
Exponential & Pow(2, 3) & 8 \\&&\\
Logarithm & Log(2, 8) & 3 \\&&\\
Timezone Converter &  TimezoneConverter("2022-01-02 22:00:00", "Asia/Shanghai", "America/New\_York")& 2022-01-02 09:00:00 \\&&\\
Sleep & Sleep(20) & Sleep for 20 seconds (\textit{Mock Response})\\&&\\
Movement Controller & RobotMove(0.3) & Robot is moving forward for 0.3 meters (\textit{Mock Response}) \\
\bottomrule
\end{tabular}
\caption{Examples of API Calls and Outputs for Each Tool}
\label{table:apicall}
\end{table*}
\end{CJK*}

\begin{CJK*}{UTF8}{gbsn}
\begin{table*}[ht]
\begin{tabular}{p{0.2\textwidth}p{0.8\textwidth}}
\toprule
\begin{tabular}{p{0.2\textwidth}}\textbf{Task}\end{tabular} & \begin{tabular}{p{0.8\textwidth}}\textbf{Demonstration} \end{tabular}\\\midrule

\begin{tabular}{p{0.2\textwidth}}Math\end{tabular} & 
\begin{tabular}{p{0.8\textwidth}} \tt You are the Calculator tool. Your task is to answer the questions that contain numbers and require arithmetic operations, including addition, subtraction, multiplication, division. Here are some examples: \\
\tt Input: There were 86 pineapples in a store. The owner sold 48 pineapples. 9 of the remaining pineapples were rotten and thrown away. How many fresh pineapples are left?\\
\tt Output: There are total 86 pineapples. 48 pineapples are sold out, so there are 86 - 48 pineapples now. 9 of the remaining are thrown away, so there are 86 - 48 - 9 pineapples. That is 29 pineapples.\end{tabular} \\&\\

\begin{tabular}{p{0.2\textwidth}}Commonsense Reasoning \end{tabular} & 
\begin{tabular}{p{0.8\textwidth}}\tt You are the Question Answering tool that answers questions by reasoning and commonsense knowledge. Here are some examples:\\
\tt Input: The women met for coffee. What was the cause of this? A: The cafe reopened in a new location. B: They wanted to catch up with each other.\\
\tt Output: Considering the options, the more likely cause for the women meeting for coffee would be B: They wanted to catch up with each other. Meeting for coffee is often chosen as a way to have a relaxed and informal conversation, providing an opportunity for friends or acquaintances to reconnect and share updates about their lives.\end{tabular} \\&\\

\begin{tabular}{p{0.2\textwidth}} Open-domain Question Answering \end{tabular} & 
\begin{tabular}{p{0.8\textwidth}}\tt You are the Wikipedia Search tool that is to look up information from Wikipedia that is necessary to answer the question. Here are some examples:\\
\tt Input: The colors on the flag of Ghana have the following meanings: green for forests, and gold for mineral wealth. What is the meaning of red?\\
\tt Output: The color Red commemorates those who died or worked for the country's independence. \end{tabular} \\&\\

\begin{tabular}{p{0.2\textwidth}}Machine Translation\end{tabular} & 
\begin{tabular}{p{0.8\textwidth}}\tt You are the Machine Translation tool that is used for translating text from one language to another. Here are some examples: \\
\tt Input: How do I ask Japanese students if they had their dinner yet? \\
\tt Output: 晩ご飯をもう食べましたか。\end{tabular} \\ \bottomrule

\end{tabular}
\caption{Example of Various Task Demonstrations for Few-Shot Baselines}
\label{table:few-shot demonstrations}
\end{table*}
\end{CJK*}

\begin{table*}[t]
\footnotesize
\centering
\resizebox{.99\textwidth}{!}{
\begin{tabular}{p{0.1\textwidth}p{0.2\textwidth}p{0.65\textwidth}}
\toprule
\begin{tabular}{p{0.1\textwidth}}\textbf{Tool}\end{tabular} & \begin{tabular}{p{0.2\textwidth}}\textbf{Description}\end{tabular}& \begin{tabular}{p{0.65\textwidth}}\textbf{Few-Shot Prompt} \end{tabular}\\\midrule

\begin{tabular}{p{0.1\textwidth}}Calculator\end{tabular} & 
\begin{tabular}{p{0.2\textwidth}}
\tt Calculator API is used for answering questions that contain numbers and require arithmetic operations, including addition, subtraction, multiplication, division. \end{tabular} & 
\begin{tabular}{p{0.65\textwidth}} \tt Calculator API is used for solving questions that require arithmetic operations, including addition, subtraction, multiplication, division. You task is to rephrase the question prepended by the special token \textless{}Q\textgreater and generate Calculator API call prepended by \textless{}API\textgreater for solving that question. You can call the API by writing "{[}Calculator(formula){]}" where "formula" is the arithmetical formula you want to solve. Here are some examples of Calculator API calls: \\ \tt Input: There were 86 pineapples in a store. The owner sold 48 pineapples. 9 of the remaining pineapples were rotten and thrown away. How many fresh pineapples are left?
\\ \tt Output: \textless{}Q\textgreater There are total 86 pineapples. 48 pineapples are sold out, so there are 86 - 48 pineapples now. 9 of the remaining are thrown away, so there are 86 - 48 - 9 pineapples. \textless{}API\textgreater {[}Calculator(86 - 48 - 9){]}.
\end{tabular} \\&&\\

\begin{tabular}{p{0.1\textwidth}}Question Answering\end{tabular} & 
\begin{tabular}{p{0.2\textwidth}}\tt Question Answering API answers questions by reasoning and commonsense knowledge. \end{tabular} & 
\begin{tabular}{p{0.65\textwidth}}\tt Question Answering API answers questions by reasoning and commonsense knowledge. You task is to rephrase the question prepended by the special token \textless{}Q\textgreater and generate QA API call prepended by \textless{}API\textgreater for solving that question. Here are some examples of API calls: You can call the API by writing "{[}QA(question){]}" where "question" is the question you want to ask. Here are some examples of QA API calls: \\ \tt Input: What do people want to acquire from opening business? A: home B: wealth C: bankruptcy D: get rich
\\ \tt Output: \textless{}Q\textgreater   What do people want to acquire from opening business? A: home B: wealth C: bankruptcy D: get rich \textless{}API\textgreater  {[}QA("What do people want to acquire from opening business? A: home B: wealth C: bankruptcy D: get rich"){]}.\end{tabular}\\&&\\

\begin{tabular}{p{0.1\textwidth}}Wiki Search\end{tabular} & 
\begin{tabular}{p{0.2\textwidth}}\tt Wikipedia Search API is to look up information from Wikipedia that is necessary to answer the question. \end{tabular} & 
\begin{tabular}{p{0.65\textwidth}}\tt Wikipedia Search API is to look up information from Wikipedia that is necessary to answer the question. You task is to rephrase the question prepended by the special token \textless{}Q\textgreater and generate Wikipedia Search API call prepended by \textless{}API\textgreater for solving that question. You can do so by writing "{[}WikiSearch(term){]}" where "term" is the search term you want to look up. Here are some examples of WikiSearch API calls:\\ \tt Input: The colors on the flag of Ghana have the following meanings: green for forests, and gold for mineral wealth. What is the meaning of red?\\ \tt Output: \textless{}Q\textgreater Ghana flag green means forests, Ghana flag gold means mineral wealth, what is the the meaning of Ghana flag red? \textless{}API\textgreater {[}WikiSearch("Ghana flag red meaning"){]}.\end{tabular}\\&&\\

\begin{tabular}{p{0.1\textwidth}}Machine Translation\end{tabular} & 
\begin{tabular}{p{0.2\textwidth}}\tt Machine Translation API is used for translating text from one language to another.\end{tabular} & 
\begin{tabular}{p{0.65\textwidth}}\tt Machine Translation API is used for translating text from one language to another. You task is to rephrase the question prepended by the special token \textless{}Q\textgreater and generate MT API call prepended by \textless{}API\textgreater for solving that question. You can do so by writing "{[}MT(text, target\_language){]}" where "text" is the text to be translated and "target\_language" is the language to translate to. Here are some examples of MT API calls:\\ \tt Input: How do I ask Japanese students if they had their dinner yet?\\ \tt Output: \textless{}Q\textgreater Translate "Did you have dinner yet" in Japanese \textless{}API\textgreater {[}MT("Did you have dinner yet?", "ja"){]}.\end{tabular}\\ \bottomrule
\end{tabular}
}
\caption{Descriptions and Usage Prompts of Four Basic Tools}
\label{table:Prompt}
\end{table*}

\end{document}